\documentclass[lettersize,journal]{IEEEtran}
\usepackage{amsmath,amsfonts}
\usepackage{algorithmic}
\usepackage{algorithm}
\usepackage{array}
\usepackage{bbding}
\usepackage{bm}
\usepackage[caption=false,font=normalsize,labelfont=sf,textfont=sf]{subfig}
\usepackage{textcomp}
\usepackage{stfloats}
\usepackage{url}
\usepackage{verbatim}
\usepackage{graphicx}
\usepackage{cite}
\usepackage{booktabs}
\usepackage{multirow}
\usepackage{hyperref}
\begin{document}

\title{Generative Cognitive Diagnosis}

\author{\textbf{Jiatong Li$^1$, Qi Liu$^1$, Mengxiao Zhu$^1$}\\
$^1$University of Science and Technology of China \\ cslijt@mail.ustc.edu.cn, \{qiliuql, mxzhu\}@ustc.edu.cn }



\maketitle

\begin{abstract}
Cognitive diagnosis (CD) models latent cognitive states of human learners by analyzing their response patterns on diagnostic tests, serving as a crucial machine learning technique for educational assessment and evaluation. Traditional cognitive diagnosis models typically follow a transductive prediction paradigm that optimizes parameters to fit response scores and extract learner abilities. These approaches face significant limitations as they cannot perform instant diagnosis for new learners without computationally expensive retraining and produce diagnostic outputs with limited reliability. In this study, we introduces a novel generative diagnosis paradigm that fundamentally shifts CD from predictive to generative modeling, enabling inductive inference of cognitive states without parameter re-optimization. We propose two simple yet effective instantiations of this paradigm: Generative Item Response Theory (G-IRT) and Generative Neural Cognitive Diagnosis Model (G-NCDM), which achieve excellent performance improvements over traditional methods. The generative approach disentangles cognitive state inference from response prediction through a well-designed generation process that incorporates identifiability and monotonicity conditions. Extensive experiments on real-world datasets demonstrate the effectiveness of our methodology in addressing scalability and reliability challenges, especially $\times 100$ speedup for the diagnosis of new learners. Our framework opens new avenues for cognitive diagnosis applications in artificial intelligence, particularly for intelligent model evaluation and intelligent education systems. The code is available at \url{https://github.com/CSLiJT/Generative-CD.git}.
\end{abstract}


\section{Introduction}

\begin{figure*}[t]
    \centering
    \includegraphics[width=\linewidth]{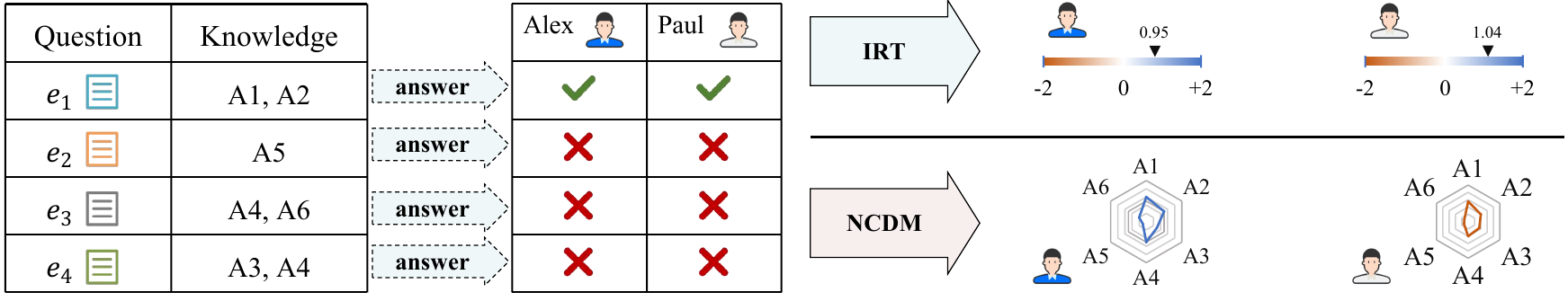}
    \caption{An overview of the cognitive diagnosis task. Given question data and response data, diagnostic models optimize parameters to fit response scores. User parameters are presented as diagnostic outputs.}
    \label{fig:cognitive-diagnosis-overview}
\end{figure*}

\par Cognitive diagnosis (CD) is a significant task in machine learning, which aims to model latent and complex cognitive states of human learners via mining their response data on diagnostic tests. As a general evaluation technique stemming from psychometrics, CD has been widely applied in various evaluation scenarios, such as student knowledge proficiency modeling, computerized adaptive testing and large language model evaluation. Figure \ref{fig:cognitive-diagnosis-overview} presents two typical cognitive diagnosis models (CDMs), the Item Response Theory (IRT) model and the Neural Cognitive Diagnosis Model (NeuralCDM or NCDM). Given question data and response data, CDMs utilize parameter estimation algorithms to fit response scores and optimize learner \& item parameters. Learner parameters are then presented as diagnostic outputs.  The IRT models learner ability as a single variable, while NCDM models learner ability as knowledge mastery degrees.
 Compared to explicit evaluation techniques such as the Classical Testing Theory (CTT), CD enables deeper understanding of the cognition construction of test-takers, providing a clearer guidance of the ability development of test-takers. Therefore, how to design accurate, trustworthy and explainable CDMs has become a significant question thesedays.
\par \textbf{Challenges}. Since the very beginning of the development of cognitive diagnosis, the design of CDMs has focused on appropriately modeling latent cognitive states via accurate fitting of response data. Researchers emphasized on designing appropriate interaction function (or, item response function in Item Response Theory) for CDMs to predict response data as accurate as possible. For instance, the Item Response Theory (IRT) uses a logisitc function to predict response data given the random variable representation of learner traits and item attributes. In this study, this paradigm is called the \textit{transductive prediction paradigm}. However, the transductive pediction paradigm of cognitive diagnosis confronts several inevitable challenges:
\begin{itemize}
    \item \textbf{Unable to diagnose instantly for new-coming learners}. When new learners comes for cognitive diagnosis, the CDM has to be retrained from scratch to update the cognitive states. This is very resource-consuming and can lead to the inconsistency for existing learner cognitive states before and after retraining.
    
    \item \textbf{Diagnostic outputs are less reliable}. The diagnostic results of CDMs are less reliable for learners and teachers, in terms of identifiability and explainability. These results are usually not identifiable because of inevitable randomness in parameter optimization. In addition, the psychometric explainability of these results are largely limited by the interaction function and data distribution.
\end{itemize}

\par \textbf{Motivation \& Contribution}. In this study, we aim to propel the shift of CD from predictive cognitive diagnosis to generative cognitive diagnosis. Inspired by generative models in machine learning, we propose a novel generative diagnosis paradigm for cognitive diagnosis. In the new paradigm, the cognitive state inference is entirely disentangled from response data prediction via a generation process. As a result, the transductive prediction paradigm is a module of the generative diagnosis paradigm. The new paradigm effectively addresses the three challenges above. For the incremental inference barrier, the new paradigm enables inductive cognitive state inference without parameter re-optimization. %
For the diagnosis reliability gap, well-designed generation process with the identifiability condition and the monotonicity condition makes diagnostic results more reliable. Next, we propose two instantiation of the generative diagnosis paradigm, the Generative Item Response Theory (G-IRT) and the Identifiable Cognitive Diagnosis Model (ID-CDM). We demonstrate the effectiveness of the proposed methods via extensive experiments on real-world cognitive diagnosis datasets. We further discussed the utility of generative cognitive diagnosis in pioneering research of artificial intelligence, such as LLM evaluation and intelligent education systems.
\section{Related Work}

\subsection{Cognitive Diagnosis}
\par Traditional cognitive diagnosis models (CDMs) are based on the transductive cognitive  paradigm. For instance, Deterministic Input, Noisy `And' gate model (DINA) \cite{Torre2009} is a discrete CDM that assumes knowledge mastery levels are binary, and utilizes a logistic-like interaction function to predict response scores from learner traits and question parameters. Item Response Theory (IRT) \cite{Fischer1995, Brzezinska2020} is a continuous CDM. In the two-parameter IRT (2PL-IRT) \cite{Fischer1995}, a learner $i$'s ability is modeled as a scalar $\theta_i$, while a question $j$ is represented by its discrimination $a_j$ and difficulty $b_j$. Then, the response score given the learner's ability and the question parameter is modeled as $P(r_{ij}=1|\theta_i,a_j,b_j)=\frac{1}{1+\exp\{-a_j(\theta_i-b_j)\}}$, where $r_{ij}$ denotes the response score. Learner abilities and question parameters are estimated by parameter optimization methods, such as full Bayesian statistical inference with MCMC sampling \cite{Gelfand1990, Hastings1970} or variational inference \cite{Wu2020}. Multidimensional Item Response Theory (MIRT) \cite{Reckase2009} further extends learner abilities and question difficulties to multidimensional cases, while the interaction function is still logistic-like. So far, deep learning techniques \cite{Yeung2019, WangF2022} have also been widely applied to CD to reach a more accurate diagnosis. For instance, NCDM \cite{WangF2022} leverages a three-layer positive full-connection neural network to capture the complex interaction between learners and questions. Along this line, recent advancements in cognitive diagnosis such as ECD~\cite{Zhou0WWHT0CM2021}, RCD~\cite{gao2021rcd} and HierCDF~\cite{Li2022} utilize neural networks to capture cognitive state information from non-behavioral data, such as learner context-aware features and knowledge dependency graphs. However, these methods still depend on transductive score prediction, which suffers from unidentifiable diagnosis and low efficiency in instant diagnosis for new learners.

\subsection{Generative Model}
\par Generative models have demonstrated remarkable success across diverse domains by learning to model the underlying data distribution and generate new samples accordingly. Traditional generative approaches, such as Variational Autoencoders (VAEs) \cite{kingma2014vae} and Generative Adversarial Networks (GANs) \cite{goodfellow2020gan}, establish the foundation for probabilistic generation through latent variable modeling and adversarial training respectively. In computer vision, diffusion models \cite{ho2020ddpm, dhariwal2020diffusion} have recently achieved state-of-the-art performance in image generation by modeling the gradual denoising process, where the generation procedure is formulated as $p_\theta(x_{t-1}|x_t) = \mathcal{N}(x_{t-1}; \mu_\theta(x_t, t), \sigma_t^2 I)$, enabling high-quality synthesis through iterative refinement. In natural language processing, encoder-decoder architectures \cite{Sutskever2014} and transformer-based models like BERT \cite{devlin2019bert} have revolutionized text generation and understanding, where masked language modeling allows bidirectional context encoding through $P(x_i|\text{mask}) = \text{softmax}(W \cdot h_i + b)$, with $h_i$ representing the contextualized hidden state. For recommender systems, generative approaches such as U-AutoRec \cite{Sedhain2015} and Collaborative Denoising Auto-Encoders (CDAE) \cite{WuDZE2016} model user-item interactions by learning to reconstruct rating matrices, where CDAE incorporates user-specific noise to capture personalized preferences through $\hat{r}_u = f(W_2 \cdot g(W_1 \cdot r_u + V \cdot n_u + b_1) + b_2)$, with $n_u$ denoting user-specific corruption. Despite the natural alignment between generative modeling and cognitive diagnosis—where both aim to model latent cognitive states and generate plausible response patterns—the generative paradigm remains largely unexplored in the cognitive diagnosis community, presenting a significant opportunity for advancing diagnostic accuracy and interpretability.

\section{Generative Diagnosis Paradigm}
\subsection{Preliminaries}
\par In cognitive diagnosis, we use $S$ to represent the learner set, $E$ to represent the item set, and $D=\{(s_i, e_j, y_{ij}) | s_i\in S, e_j\in E, y_{ij}\in \mathbb{R}\}$ to represent the response score set, respectively. The $y_{ij}$ represents the response score of $s_i$ on $e_j$. In the classical setting of cognitive diagnosis, response scores are binary. This is common in objective tests items such as multiple-choice questions. For subjective test items, response scores are usually in an ordinal scale.
\par Meanwhile, the learner latent trait set is represented by $\Theta$. The item attribute set is represented by $\Psi$. One basic assumption in cognitive diagnosis is that $\Theta$ and $\Psi$ decide response scores. Unfortunately, $\Theta$ and $\Psi$ are unobservable. So cognitive diagnosis models first assumes the interaction between learner latent traits and item attributes via a well-defined \textit{Item Response Function} (IRF), then finds the optimal latent variables by fitting response score data. Formally, the cognitive diagnosis task is defined as follows:
\newtheorem{definition}{Definition}[section]
\begin{definition}\label{def:cognitive-diagnosis}
    \textbf{Cognitive diagnosis}. Given the learner set $S$, item set $E$, response score set $D$, assuming  the item response function as $f_{\omega}$ and the distance metric as $\mathcal{L}$, the goal of cognitive diagnosis is to find optimal learner latent traits $\Theta^*$ and item attributes $\Psi^*$, such that
    \begin{equation}
        \Theta^*, \Psi^* = \arg\min_{\Theta,\Psi,\omega} \mathbf{E}_{(s_i,e_j,y_{ij})\sim D}\left[\mathcal{L}\left(y_{ij}, f_{\omega}(\theta_i;\psi_j)\right)\right]
    \end{equation}
\end{definition}

\par For instance, in two-parameter Item Response Theory (2PL-IRT) models, each learner trait $\theta_i$ is a random variable representing its overall ability, while each item attribute $\psi_j=(a_j,b_j)$ is a pair of random variables representing its discrimination level $a_j$ and difficulty level $b_j$. The item response function is 
\begin{equation}
    f_{\omega}^{(IRT)}(\theta_i;\psi_j) = \frac{1}{1+\exp(-a_j(\theta_i-b_j))}.
\end{equation}
\par In this case, the IRF output is viewed as the probability of a successful binary response score (i.e., $y_{ij} = 1$). Therefore, the distrance metric $\mathcal{L}^{(IRT)}$ is the negative log-likelihood function. That is, 
\begin{equation}
    \mathcal{L}^{(IRT)}(y, \hat{y}) = -\left[y\log \hat{y} + (1-y)\log(1-\hat{y})\right].
\end{equation}
\par In previous research, the estimation of latent variables of cognitive diagnosis depends directly-applied parameter estimation algorithms, such as the Monte Carlo Markov Chain (MCMC) for distribution estimation of cognitive states or Gradient Descent (GD) for point estimation of cognitive states. 

\subsection{Reliability Requirements of Diagnostic Results}
\begin{definition}\label{def:identifiability}
  \textbf{Identifiability in cognitive diagnosis}. Let $B=\{\Theta, \Psi\}$ be the set of diagnostic results, and let $\{f_{R}(\bm{\theta};\bm{\psi}):\Theta\times\Psi\to\{0,1\}|\bm{\theta}\in\Theta,\bm{\psi}\in\Psi\}$ be the set of response function which generate response data given diagnostic results. Furthermore, let $\bm{y}_i^{(s)} = f_R(\bm{\theta}_i;\cdot)$ be the response data distribution of learner $s_i$ with trait $\theta_i$. Let $\bm{y}_k^{(e)} = f_R(\cdot;\bm{\psi}_k)$ be the response data distribution of question $e_i$ with feature $\psi_k$. Then the set of diagnostic results is identifiable if and only if distinct diagnostic results lead to distinct distribution of response data. Specifically, the identifiability of learner traits connotes that
  \begin{equation}\label{eq:id1}
    \bm{y}_i^{(s)} = \bm{y}_j^{(s)}\rightarrow\bm{\theta}_i = \bm{\theta}_j,
  \end{equation}
  In addition, the identifiability of question parameters connotes that
  \begin{equation}\label{eq:id2}
    \bm{y}_k^{(e)}= \bm{y}_l^{(e)} \rightarrow \bm{\psi}_k=\bm{\psi}_l.
  \end{equation}
  Then a set of diagnostic results is identifiable if both Eq.\eqref{eq:id1} and Eq.\eqref{eq:id2} are satisfied.
\end{definition}

\begin{definition}\label{def:explainability}
  \textbf{Explainability in cognitive diagnosis}. The explainability of learners' diagnostic results is defined as the ability they correctly reflect learners' actual cognitive states.
\end{definition}

\par For example, if a learner has mastered the knowledge concept `\textit{Inequality}', then the component value of the diagnosed learner trait on this knowledge concept should be high so that the diagnostic result can correctly reflect the fact that the learner has mastered the knowledge concept. However, it is difficult to directly keep the explainability of diagnostic results because learners' true mastery levels are unobservable. As a result, in CD-based learner modeling, the explainability of diagnostic results is usually indirectly satisfied by the monotonicity assumption \cite{Reckase2009,WangF2022}:

\begin{definition}\label{def:mono}
  \textbf{Monotonicity assumption}. The probability of every learner correctly answering a question is monotonically increasing at any relevant component of his/her knowledge mastery level. Formally, the monotonicity assumption is equivalent to:
  \begin{equation}
    \bm{\theta_i}^{(l)} \succeq \bm{\theta_j}^{(l)} \Leftrightarrow y_{il} \geq y_{jl}, \forall s_i, s_j \in S,\,\,e_l \in E,
  \end{equation}
where $\bm{\theta_i}^{(l)} (s_i\in S, e_l\in E)$ denotes the relevant component of $s_i$'s knowledge mastery level $\bm{\theta}_i$ to question $e_l$.
\end{definition}

\par For transductive CDMs, the monotonicity assumption usually depends on the monotonicity property of the interaction function \cite{WangF2022}. For traditional CDMs such as DINA \cite{Torre2009} and IRT \cite{Brzezinska2020}, the interaction function is usually linear, thus inherently satisfying the monotonicity assumption. For deep learning-based CDMs such as NCDM \cite{WangF2022}, the weight parameter of the interaction function is limited to be non-negative to satisfy the assumption.

\subsection{Generative Diagnosis Function}
\par The key idea of generative cognitive diagnosis is to estimate cognitive states via a \textbf{generation process} rather than an optimization process. This is accomplished via a well-defined generative diagnosis function (GDF). Formally, GDF itself is represented by $g_\phi: \mathbb{R}^Z\to\Theta\times\Psi$, parameterized by $\phi$. Here $Z$ denotes the scale of the input data. GDF outputs the latent traits via a generation process from response score data:
\begin{equation}
    \bm{\theta}_i, \bm{\psi}_j = g_{\phi}\left(\bm{y}_{i}^{(s)}; \bm{y}_{j}^{(e)}\right),
\end{equation}
here $y_{i}^{(s)}$ represents all response scores related to learner $s_i$, and $y_{j}^{(e)})$ represents all response scores related to item $e_j$. Overall, the GDF includes two steps: 1) Data aggregation and 2) Feature generation. The first step aggregates the sparse, huge response score data into a denser and smaller representation for the next step. The second step then extracts latent trait and item feature information from the condensed representation to generate diagnostic outputs. We will show how this process works using specific instantiation of the generative diagnosis paradigm in Section [3] and [4]. 

\par By shifting cognitive state estimation from parameter estimation to diagnostic generation, the goal of cognitive diagnosis shifts from finding the optimal $\Theta$ and $\Psi$ to finding the optimal GDF parameter $\phi$. \textbf{Moreover, the cognitive state diagnosis process is disentangled from the score prediction process in the inference stage.} This leads to many significant advantages of the generative diagnosis paradigm compared to the transductive prediction paradigm:
\begin{itemize}
    \item \textbf{Efficacy in diagnosing new-coming learners}. When new learners comes, their cognitive state estimation could be instantly obtained by directly inputting their response scores to the GDF and running the generation process, without retraining the whole model.
    \item \textbf{Reliability in latent trait estimation}. The latent trait estimation of GDF is highly controllable because one can conveniently apply parameter mediation to the GDF. This leads to a comprehensive understanding of the explainability and causality between response data and cognitive states generated by GDF.
\end{itemize}
\par It should be noted that these advantages mentioned above are exclusive to the new paradigm. For incremental inference, the transductive prediction paradigm needs retraining the whole model to avoid parameter overfitting, which is resource-consuming when the incoming learners are frequent. For data utilization, the transductive prediction paradigm has to change the IRF, which is complex and could destroy the well-defined feature interaction relationship. For latent trait estimation, the estimation output is less controllable since it is instantly updated by the optimization algorithm during training. Applying intervention on estimation could affect the training of the whole model.

\begin{figure*}[t]
    \centering
    \includegraphics[width=\linewidth]{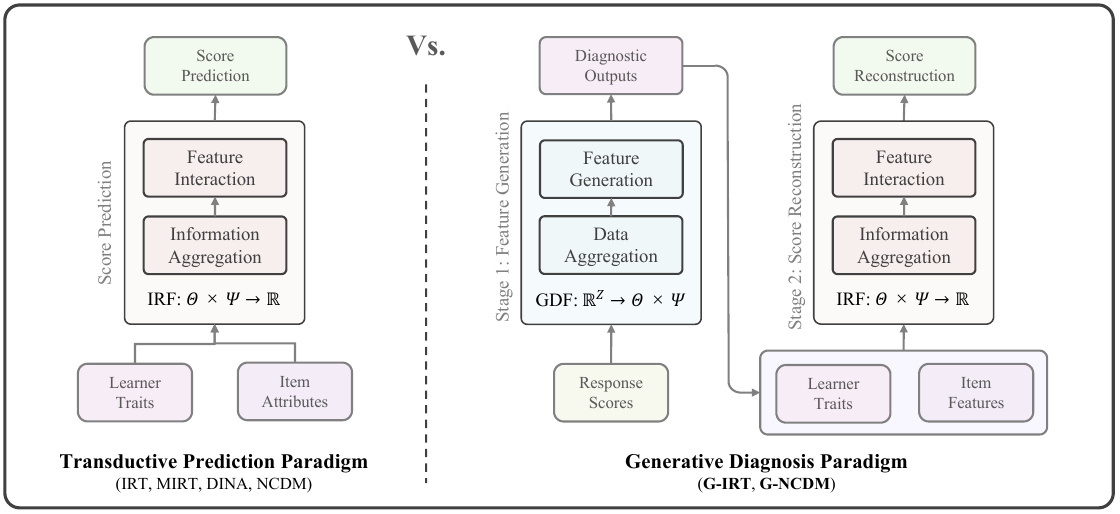}
    \caption{An overview of the classic transductive prediction paradigm (left) and the new generative diagnosis paradigm (right).}
    \label{fig:paradigm-overview}
\end{figure*}

\subsection{Item Response Function}
The item response function (IRF) defines the interaction relationship between learner traits and item features, which is mathematically identical to the transductive prediction paradigm. Formally, IRF is represented by $f_\omega:\Theta\times\Psi\to\mathbb{R}$, parameterized by $\omega$. IRF then outputs the prediction of response score from latent traits:
\begin{equation}
    \hat{y}_{ij} = f_{\omega} (\bm{\theta}_i; \bm{\psi}_j).
\end{equation}
\par Differently, the IRF plays an assistant role in helping the learning of the generative diagnosis function via response score reconstruction, rather than dominating the latent trait estimation.

\subsection{Optimization Objective}
\par Given the generative diagnosis function and the item response function above, the training of a generative cognitive diagnosis model becomes a parameter optimization task, similar to the classical cognitive diagnosis in Definition~\ref{def:cognitive-diagnosis}. Here we give a formal definition of the optimiztion objective.

\begin{definition}[Optimization Objective of Generative Diagnosis Paradigm]
    Given the generative diagnosis function $g_\phi$, the item response function $f_{\omega}$, the learner set $S$,  the item set $E$ and the response score set $D$, the goal of generative diagnosis pradigm is to find optimal function parameter $\phi^*$ and $\omega^*$ such that
    \begin{equation}
        \phi^*,\omega^* = \arg\min_{\phi,\omega}\mathbf{E}_{(s_i,e_j,y_{ij})\sim D}\left[\mathcal{L}\left(y_{ij}, f_{\omega}(\bm{\theta}_i;\bm{\psi}_j)\right)\right],
    \end{equation}
    here $\bm{\theta}_i$ and $\bm{\psi}_j$ are obtained by
    \begin{equation}
        \bm{\theta}_i, \bm{\psi}_j = g_{\phi}(\bm{y}_{i}^{(s)}; \bm{y}_{j}^{(e)}).
    \end{equation}
\end{definition}
\par Evidently, the optimization objective of generative diagnosis paradigm is similar to Definition~\ref{def:cognitive-diagnosis}, except that optimization objective itself shifts from latent traits to GDF parameters.
\section{G-IRT: Generative Item Response Theory}
\par As introduced above, IRT and NeuralCDM are two of the most typical CDMs in the literature. In this section, we introduce the Generative Item Response Theory (G-IRT), which could obtain learner and item traits via feature generation without changing the original IRT structure.

\begin{figure*}
    \centering
    \includegraphics[width=\linewidth]{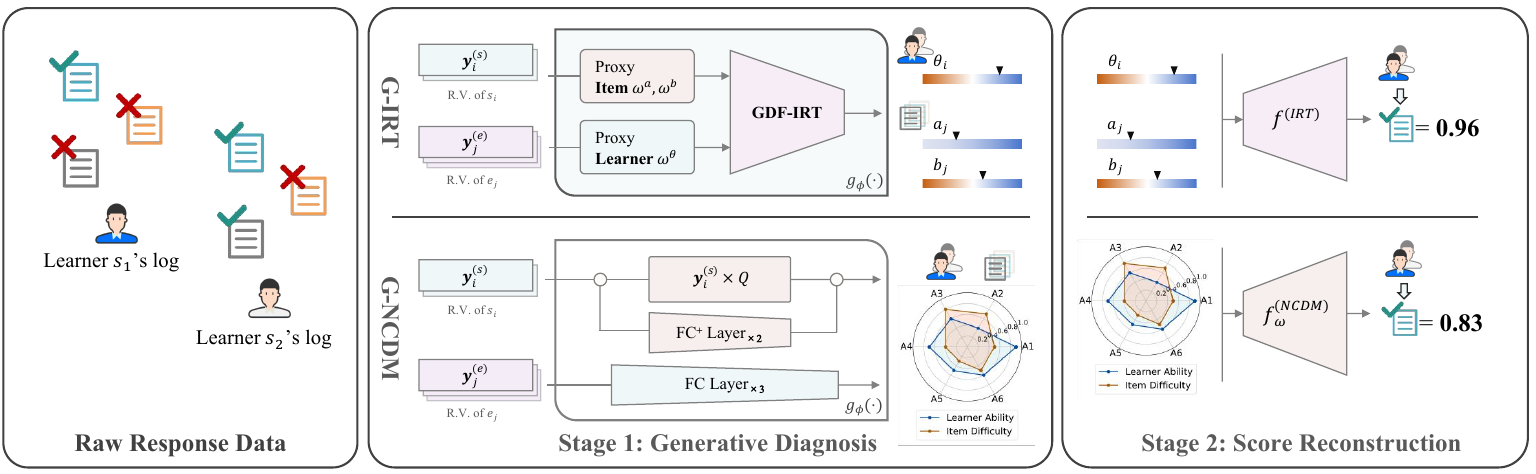}
    \caption{An overview of models under the Generative Diagnosis paradigm. The upper parts present G-IRT. The lower parts present G-NCDM. R.V. denotes response score vectors.}
    \label{fig:model-overview}
\end{figure*}
\subsection{Challenges in Latent Trait Estimation of IRT}
\par IRT is a classical CDM that utilizes machine learning techniques to optimize learner and item parameters. Challenges in the estimation of latent traits includes \textit{controllablity and efficiency}. The \textit{controllability} challenge lies in the interdependence of latent traits during the estimation process. To jointly estimate respondent and item parameters, various estimation algorithms such as MCMC, variational inference (VI) and gradient descent (GD) were applied to the estimation of IRT parameters. However, it is non-trivial to interpret and control the relationship between latent traits and specific score distributions because parameters are deductively learned. The \textit{efficiency} challenge lies in the incremental estimation of respondent traits, which is common in the deployment phase of an IRT model. Although it might take only 10 seconds to re-estimate the IRT give one new respondent, the accumulated time cost could be large when the number of new respondents could be very large.


\subsection{The Generative Diagnosis Function of G-IRT}
\par Here we demonstrate the GDF of G-IRT using IRT-2PL. In the traditional 2PL-IRT, the item response function is defined as 
\begin{equation}
    P(y_{ij}|\theta_i,a_j,b_j) = \frac{1}{1+\exp(-a_j(\theta_i-b_j))}.
\end{equation}

Therefore, if we know two of the three latent traits $\theta_i$, $a_j$ and $b_j$, we could estimate the last one with the estimated correct probability $P(y_{ij}|\theta_i,a_j,b_j)$. This means the estimation of each parameter could be decided by 

\begin{equation}\label{eq:irt-theta}
    \theta_i|_{a_j,b_j} = b_j + \frac{\sigma^{-1}(P_{ij})}{a_{j}},
\end{equation}

\begin{equation}\label{eq:irt-a}
    a_j|_{b_j,\theta_i} = \frac{\sigma^{-1}(P_{ij})}{\theta_i-b_j},
\end{equation}

\begin{equation}\label{eq:irt-b}
    b_j|_{\theta_i,a_j} = \theta_i - \frac{\sigma^{-1}(P_{ij})}{a_{j}}.
\end{equation}

\par However, this estimation is impractical for traditional transductive prediction paradigm because the three parameters are unobservable and we only have access to response scores $R_{ij}$ rather than the correct probability. Alternatively, latent parameter estimation algorithms such as Expectation-Maximum (EM) and MCMC are widely used for IRT. However, this estimation can be easily accomplished within the generative diagnosis paradigm, since we can implement it via the generative diagnosis function. The key idea is to replace unobserved prior parameters with \textbf{proxy parameters}. Proxy parameters play the same role as in the right side of Eq. \ref{eq:irt-theta}, \ref{eq:irt-a} and \ref{eq:irt-b}. However, they only serve for parameter estimation and do not represent diagnostic learner/item traits. In G-IRT, we first define the latent trait estimation conditioned on proxy parameters as 

\begin{equation}\label{eq:girt-theta-single}
    \theta_{i}(\omega^{a}_j,\omega^{b}_j) = \omega^{b}_j + \frac{\lambda R_{ij}}{\omega^{a}_j},
\end{equation}

\begin{equation}
    a_j(\omega^\theta_i,\omega_j^{b}) = \left|\frac{\lambda R_{ij}}{\omega^\theta_i - \omega_j^{b}}\right|,
\end{equation}

\begin{equation}
    b_j(\omega^{\theta}_i,\omega^{a}_j) = \omega^{\theta}_i - \frac{\lambda R_{ij}}{\omega^{a}_j}.
\end{equation}
\par Here $\lambda\in \mathbb{R}_+$ is a hyperparameter. In this definition, $\sigma^{-1}(P_{ij})$ is approximated by $\lambda R_{ij}$. The $\omega^\theta_{\cdot}$, $\omega^{a}_{\cdot}$ and $\omega^{b}_{\cdot}$ are the \textit{proxy parameters} for estimating latent traits. Now the problem is, for each observed $R_{ij}\neq 0$, we can calculate an estimated value, namely $\theta_{i}(\omega^{a}_j,\omega^{b}_j)$. We aim find a general $\theta_i$ that can minize its distance between every $\theta^{\omega}_{ij}, j=1,2,\ldots,M$. Our goal is to

\begin{equation}
    \hat{\theta_i} = \arg\min_{\theta} \sum_{j: R_{ij}\neq 0} (\theta-\theta_{i}(\omega^{a}_j,\omega^{b}_j))^2.
\end{equation}

We can establish similar goal for $\hat{a}_j$ and $\hat{b}_j$.  The solution of them is

\begin{equation}\label{eq:girt-theta-overall}
    \theta_i = \frac{1}{Z_i^{(s)}}\sum_{j: R_{ij}\neq 0}\theta_{i}(\omega^{a}_j,\omega^{b}_j),
\end{equation}
here $Z_i^{(s)} = \sum_{j=1}^{M}\mathbb{I}(R_{ij}\neq 0)$. Similarly, for item parameters, we have
\begin{equation}
    a_j = \frac{1}{Z_j^{(e)}}\sum_{i: R_{ij}\neq 0}a_j(\omega^\theta_i,\omega_j^{b}),
\end{equation}
\begin{equation}
    b_j = \frac{1}{Z_j^{(e)}}\sum_{i: R_{ij}\neq 0} b_j(\omega^{\theta}_i,\omega^{a}_j)
\end{equation}

To transform these definition to the form of GDF, we first replace the response log $R_{ij}\in\{-1,0,1\}$ with $y_{ij}\in\{0,1\}$. Since $R_{ij} = 1$ denotes correct response, $R_{ij} = 0$ denotes no response and $R_{ij} = -1$ denotes incorrect response, we have $R_{ij} = 2y_{ij}-1, \forall R_{ij}\neq 0$, and $R_{ij}\neq 0$ means $y_{ij}\notin y^{(s)}_i$ and $y_{ij}\notin y^{(e)}_j$. As a result, we can define the generative diagnosis function of G-IRT as follows:

\begin{equation}
    g_{\phi}(y_{i}^{(s)}; y_{j}^{(e)}) \equiv
    \left\{\begin{aligned}
        \theta_i &= \frac{1}{|\bm{y}_i^{(s)}|}\sum_{j:y_{ij}\in y_i^{(s)}} \left[\omega^{b}_j + \frac{\lambda (2y_{ij}-1)}{\omega^{a}_j}\right]\\
        a_j &= \frac{1}{|\bm{y}_{j}^{(e)}|}\sum_{i: y_{ij}\in y_{j}^{(e)}}\left|\frac{\lambda (2y_{ij}-1)}{\omega^\theta_i - \omega_j^{b}}\right|\\
        b_j &= \frac{1}{|\bm{y}_{j}^{(e)}|}\sum_{i: y_{ij}\in y_{j}^{(e)}} \left[\omega^{\theta}_i - \frac{\lambda R_{ij}}{\omega^{a}_j}\right]
    \end{aligned}\right.
\end{equation}

Here $\phi = (\omega^\theta,\omega^a,\omega^b)$ is not only the proxy parameter set, but also the optimization target of the GDF.

\renewcommand{\algorithmiccomment}[1]{\hfill $\triangleright$ #1}
\begin{algorithm}
\caption{The training of G-IRT}\label{alg:training-girt}
\textbf{Input}: $D=\{(s_i, e_j, y_{ij}) | s_i\in S, e_j\in E, y_{ij}\in \mathbb{R}\}$, the training dataset \\
\textbf{Output}: $g_\phi$, the GDF of a trained G-IRT

\begin{algorithmic}[1]
\STATE Initialize $g_\phi$;
\FOR{epoch in $1\ldots T$}
\FOR{$(s_i,e_j,y_{ij})\in D$}
\STATE $\bm{y}_i^{(s)}\leftarrow[y_{ij'}|j': (s_i,e_{j'},y_{ij'})\in D]$; \\
\COMMENT{Obtain the learner response vector for $s_i$}
\STATE $\bm{y}_j^{(e)}\leftarrow[y_{i'j}|i': (s_{i'},e_{j},y_{i'j})\in D$; \\
\COMMENT{Obtain the item response vector for $e_j$}
\STATE $\theta_i, a_j, b_j \leftarrow g_{\phi}\left(\bm{y}_{i}^{(s)};\bm{y}_{j}^{(e)}\right)$; \\
\COMMENT{Stage 1: Feature generation for learner and item}
\STATE $\hat{y}_{ij} \leftarrow f^{(IRT)} (\theta_i, a_j, b_j)$; \\
\COMMENT{Stage 2: Score reconstruction using IRT}
\ENDFOR
\STATE $ g_{\phi}\leftarrow \text{update}\left(\mathcal{{L}}(y, \hat{y}), g_{\phi}\right)$; \\
\COMMENT{Update GDF using gradient descent}
\ENDFOR
\STATE \textbf{return} $g_{\phi}$;
\end{algorithmic}
\end{algorithm}

\begin{algorithm}
\caption{Instant diagnosis using G-IRT}\label{alg:diagnosis-girt}
\textbf{Input}: $g_{\phi}$, the GDF of a trained G-IRT; $D_i$, the response scores of a new-coming learner $s_i$\\
\textbf{Output}: $\theta_i$, the learner ability value

\begin{algorithmic}[1]
    \STATE $\bm{y}_i^{(s)}\leftarrow \text{transform}(D_i)$;\COMMENT{Transform the $D_i$ into response score vector, which scores indexed by item IDs.}
    \STATE $\theta_i\leftarrow g_{\phi}^{(s)}\left(\bm{y}_i^{(s)}\right)$; \COMMENT{Generative diagnosis}
    \STATE \textbf{return} $\theta_i$;
\end{algorithmic}
\end{algorithm}

\subsection{An Analysis of G-IRT}

\subsubsection{Controllabiity of Parameter Estimation}
\par A common setting of IRT is that latent traits follow a specific but unobserved distribution. For tradition IRT models, setting appropriate prior distribution for latent traits (e.g., standard Gaussian distribution for $\theta$) could control the posterior distribution of them to control their scale. Controllable scale help learners to understand their relative ability compared to the whole learner group.
\par For G-IRT, however, the scale of latent traits is decided by the GDF parameter $\phi$ rather than their prior distribution. Here we analyze how to control the scale of $\phi$ to control the parameter scale of $\theta$, $a$ and $b$.

\par For convenience, we assume that every $R_{ij}$ is non-zero. We further assume the following hyperparameters: let $\omega_j^b\in(\alpha,\beta), j = 1,\ldots,M$; let $\omega_i^\theta\in(\alpha,\beta), i = 1,\ldots,N$; let $\omega_j^a\in(\epsilon,\zeta), j=1,\ldots,M$.

\noindent\textbf{Parameter Scale of $\theta_i$}. We start with discussing the range of respondent parameter $\theta_i$. For the upper bound of $\theta_i$, we have
\begin{equation}
    \theta_i \leq \overline{\omega_b} + \lambda\left(\frac{1}{M}\sum_{j=1}^M\frac{1}{\omega_{j}^a}\right) < \beta + \frac{\lambda}{\epsilon}.
\end{equation}
\par We aim to limit the range of $\theta_i$ within $(p,q)$. Therefore,
\begin{equation}
    \lambda \leq \epsilon(q-\beta).
\end{equation}
\par Similarly, we have
\begin{equation}
    \lambda\leq\epsilon(\alpha-p).
\end{equation}

\noindent\textbf{Parameter Scale for Cold-start Setting}. A significant advantage of G-IRT is its utility in cold-start learner modeling. Consider a new learner with its response score that has never been observed in training data. The G-IRT could directly input its response score vector to the generative diagnosis function to obtain its ability. However, the cold-start performance of G-IRT is not only decided by its training procedure, but the hyperparameter setting. Considering this, we aim to appropriately set hyperparameters of G-IRT so that it can at least handle some special cases. Specifically, we discuss a special case of all-correct response data, i.e., $R = \mathbf{1}_{N\times M}$. In this case, our anticipated generated learner ability should be always greater than item difficulty. By computing these value, we have
\begin{equation}
    \theta_i = \frac{1}{M}\sum_{j=1}^M\omega_j^b + \frac{\lambda}{M}\sum_{j=1}^M\frac{1}{\omega_j^a}
\end{equation}

\begin{equation}
    b_j = \frac{1}{N}\sum_{i=1}^N\omega_i^\theta - \lambda\frac{1}{\omega_j^a}
\end{equation}

Therefore, we have
\begin{equation}
    \theta_i - b_j = \overline{\omega_b} - \overline{\omega_\theta} + \lambda\left(\frac{1}{\omega_j^a}+\frac{1}{M}\sum_{j'=1}^M\frac{1}{\omega_{j'}^a}\right).
\end{equation}

We hope $(\theta_i-b_j)$ is greater than zero so that G-IRT has a good cold-start performance. Therefore, we limit
\begin{equation}
    \lambda > \frac{\max(\overline{\omega_b} - \overline{\omega_\theta})}{\min(\frac{1}{\omega_j^a}+\frac{1}{M}\sum_{j'=1}^M\frac{1}{\omega_{j'}^a})} \geq \frac{\zeta}{2}(\beta-\alpha).
\end{equation}

Surprisingly, the constraint for cold-start setting leads to a lower bound of $\lambda$, which jointly work with upper bounds of $\lambda$ mentioned above to decide its region.

\noindent\textbf{Make Lower and Upper Bounds Compatible}. We aim to let the lower bound and upper bound of $\lambda$ compatible to simultaneously satisfy the two properties. Therefore, we have
\begin{equation}
    \frac{\zeta}{\epsilon}\leq \min\left\{\frac{2(\alpha - p)}{\beta-\alpha}, \frac{2(q - \beta)}{\beta-\alpha}\right\}
\end{equation}

\section{G-NCDM: Generative Neural Cognitive Diagnosis Model}
\par In this section, we introduce the application of the generative diagnosis paradigm in deep learning-based CDMs. We propose the generative neural cognitive diagnosis model\footnote{G-NCDM is developed from our proposed ID-CDF~\cite{li2024idcdf}.} (G-NCDM), a simple yet effective method that captures the complex mapping between response scores and features.

\subsection{Challenges in Deep Learning-based CDMs}
\par As introduced in Figure \ref{fig:cognitive-diagnosis-overview}, existing deep learning-based CDMs model learner abilities via knowledge mastery degrees. \textbf{These CDMs mingle learner ability diagnosis and model training together, which leads to their inability in instant diagnosis and a lack of reliability.} Especially respecting reliability, deep learning-based CDMs confront with the non-identifiability problem and the explainability overfitting problem. \textit{Identifiability} plays a significant role in CD-based learner modeling and has been discussed in many works \cite{Xu2018, Xu2019identifiability}, which connotes that diagnostic results should be able to distinguish between learners with different response data distribution. That is, different learner traits should lead to different response data distributions.  \textit{Explainability} is the ability that diagnostic results truly reflect \textit{actual} cognitive states. For deep-learning based CDMs, the explainability is measured by the monotonicity between knowledge mastery degrees and real response scores. However, we notice in experiments for the first time that existing CDMs suffer from the explainability overfitting problem. That is, diagnostic results are highly explainable in observable response data for training, while less explainable in unobservable response data for testing.


\subsection{The Generative Diagnosis Function of G-NCDM}
\par The principle of the design of the GDF of G-NCDM is to satisfy the identifiability condition and the monotonicity condition while ensuring diagnosis preciseness. To this end, in ID-CDM, we adopt fully connected layers with parameter constraints to learn the diagnosis process from data, while keeping the two pivotal conditions. Specifically, the GDF is defined as follows:
\begin{equation}
    g_\phi\left(\bm{y}_i^{(s)};\bm{y}_j^{(e)}\right) \equiv \left\{ g_\phi^{(s)}\left(\bm{y}_i^{(s)}\right); g_\phi^{(e)}\left(\bm{y}_j^{(e)}\right)\right\},
\end{equation}
\begin{gather}\label{eq:gncdm-gdf}
     \bm{\theta}_i^{\text{(imp)}} = \text{FC}^{+}_{\times2}\left(2\bm{y}_i^{(s)}-1\right), \\
     \bm{\theta}_i^{\text{(exp)}} = \sigma\left(\frac{\left(2\bm{y}_i^{(s)}-1\right)\times Q}{\sqrt{K}}\right), \\
     \bm{\theta}_i = (1-\alpha)\cdot\bm{\theta}_i^{\text{(imp)}} + \alpha\cdot\bm{\theta}_i^{\text{(exp)}}.
\end{gather}
\begin{equation} 
     \bm{\psi}_j = \text{FC}_{\times3}\left(2\bm{y}_j^{(e)}-1\right).
\end{equation}
\par Here all activation functions are sigmoid. The $\text{FC}^{+}(\cdot)$ denotes fully connected layer with non-negative weight parameters, which is designed for ensuring the monotonicity between response scores and knowledge mastery degrees.

\subsection{The Item Response Function of G-NCDM}
\par In the IRF, we also adopt neural networks to learn the complex interaction between learners and questions. Specifically, we first utilize single-layer perceptrons to aggregate knowledge concept-wise diagnostic results to low-dimensional features to gain more effective representations of learners and questions. Next, we utilize fully connected layers to reconstruct response scores from aggregated representations.
\par The aggregation layer of diagnostic output is defined as follows:

\begin{gather}
    \bm{\theta}^{(\text{dense})}_{i,j} = \text{FC}^{+}(\bm{\theta}_i\odot \bm{q}_j), \\
    \bm{\psi}^{(\text{dense})}_j = \text{FC}(\bm{\psi}_j\odot \bm{q}_j), \\
\end{gather}
here $q_j$ is the $j$-th row of the expert-labeled Q-matrix \cite{Tatsuoka1983}. $q_j$ is for masking unnecessary knowledge dimensions to ensure that the optimization only targets for knowledge dimensions required by the item. for $\bm{q}_j = (q_{j,1},\ldots,q_{j,K})$, the $q_{j,l}, l=1,2,\ldots,K$ is a binary value denotes whether knowledge dimension $l$ is required by item $j$.
\par Next, aggregated representations of learner $s_i$ and question $e_j$ are input to a three-layer fully connected neural network to reconstruct correct probability, as defined as follows:
\begin{gather}
    \hat{y}_{ij} = \text{FC}_{\times 3}\left(\bm{\theta}^{(\text{dense})}_{i,j} - \bm{\psi}^{(\text{dense})}_j\right)
\end{gather}

\par Finally, neural network parameters of the GDF and the IRF are learnerbale parameters. The loss function of G-NCDM is the cross entropy loss between actual response scores and the reconstructed correct probabilities. The model can be trained via gradient descent algorithms. Specifically, the training algorithm of G-NCDM is given as follows:

\begin{algorithm}
\caption{The training of G-NCDM}\label{alg:training-gncdm}
\textbf{Input}: $D=\{(s_i, e_j, y_{ij}) | s_i\in S, e_j\in E, y_{ij}\in \mathbb{R}\}$, the training dataset; $Q=(q_{jl})_{M\times K}$, the item-knowledge Q-matrix \\
\textbf{Output}: $g_\phi$, the GDF of a trained G-NCDM

\begin{algorithmic}[1]
\STATE Initialize $g_\phi$ and $f_\omega$;
\FOR{epoch in $1\ldots T$}
\FOR{$(s_i,e_j,y_{ij})\in D$}
\STATE $\bm{y}_i^{(s)}\leftarrow[y_{ij'}|j': (s_i,e_{j'},y_{ij'})\in D]$; \\
\COMMENT{Obtain the learner response vector for $s_i$}
\STATE $\bm{y}_j^{(e)}\leftarrow[y_{i'j}|i': (s_{i'},e_{j},y_{i'j})\in D$; \\
\COMMENT{Obtain the item response vector for $e_j$}
\STATE $\bm{\theta}_i, \bm{\psi}_j\leftarrow g_{\phi}(\bm{y}_{i}^{(s)};\bm{y}_{j}^{(e)})$; \\
\COMMENT{Stage 1: Feature generation for learner and item}
\STATE $\hat{y}_{ij} \leftarrow f_\omega (\bm{\theta}_i, \bm{\psi}_j; Q)$; \\
\COMMENT{Stage 2: Score reconstruction using neural network}
\ENDFOR
\STATE $ g_{\phi}\leftarrow \text{update}\left(\mathcal{{L}}(y, \hat{y}), g_{\phi}\right)$; \\
\COMMENT{Update GDF using gradient descent}
\ENDFOR
\STATE \textbf{return} $g_{\phi}$;
\end{algorithmic}
\end{algorithm}

\begin{algorithm}
\caption{Instant diagnosis using G-NCDM}\label{alg:diagnosis-gncdm}
\textbf{Input}: $g_{\phi}$, the GDF of a trained G-NCDM; $D_i$, the response scores of a new-coming learner \\
\textbf{Output}: $\bm{\theta}_i$, the learner ability value

\begin{algorithmic}[1]
    \STATE $\bm{y}_i^{(s)}\leftarrow \text{transform}(D_i)$;\COMMENT{Transform the $D_i$ into response score vector, which scores indexed by item IDs.}
    \STATE $\bm{\theta}_i\leftarrow g_{\phi}^{(s)}(\bm{y}_i^{(s)})$; \COMMENT{Generative diagnosis}
    \STATE \textbf{return} $\bm{\theta}_i$;
\end{algorithmic}
\end{algorithm}
\section{Experiments}
\subsection{Experiment Overview}
\par We aim to answer the following research questions via experiments:
\begin{itemize}
    \item \textbf{RQ1 (Instant diagnosis for new learners)}. How is the score reconstruction performance of generative CDMs? 
    \item \textbf{RQ2 (Offline diagnosis for existing learners)}. How is the score prediction performance of generative and transductive CDMs?
    \item \textbf{RQ3 (Reliability of diagnostic outputs)}. How is the identifiability and explainability of diagnostic outputs of different CDMs?
    \item \textbf{RQ4 (Statistical features of diagnostic outputs)}. How is the statistical feature of diagnostic outputs of generative CDMs?
    \item \textbf{RQ5 (Utility of diagnostic outputs)}. How can generative CDMs be used in real-world applications?
\end{itemize}

\subsection{Experiment Setup}
\subsubsection{Datasets}
\par In this study, we evaluate the performance of G-IRT and G-NCDM on two representative intelligent education datasets, the ASSIST (ASSISTments 2009-2010 ``skill builder'') \cite{Feng2009} and the Math1 \cite{Liu2018} dataset. These datasets reflects two classical scenarios of cognitive diagnosis, including I) \textbf{Online learning platforms}, which have a large number of learners, various test items and sparse response scores, and II) \textbf{Offline ability tests}, which have a large number of learners, less but well-designed test items and dense response scores. ASSIST is a response log dataset collected from an online learning platform, with $4,163$ learners, $17,746$ items and $123$ knowledge concepts, reflecting \textit{online learner tests}. Math1 is a response log dataset collected from an offline exam, with $4,209$ learners, $20$ items and $11$ knowledge concepts,  reflecting \textit{offline ability tests}. Details of the datasets are available at Table \ref{tab:dataset-overview}. In experiments, we split datasets by $D_{train}:D_{valid}:D_{test}=70\%:10\%:20\%$.

\begin{table}[t]
\centering
\caption{A summary of experiment datasets.}
\label{tab:dataset-overview}
\begin{tabular}{l|rr}
\toprule
Statistics & ASSIST-0910 & Math-1 \\
\midrule
Scenario & Online learning platform & Offline ability test \\
\# Learners & 4,163 & 4,209 \\
\# Items & 17,746 & 20 \\
\# KCs & 123 & 11 \\
\# KCs per item & 1.19$\pm$0.47 & 3.35$\pm$1.31 \\
\# Scores & 324,572 & 84,180 \\
\# Scores per learner & 107.26$\pm$155.25 & 20.0$\pm$0.0 \\
Sparsity rate & 99.6\% & 0.0\% \\
Correct rate & 65.4\% & 42.4\% \\
\bottomrule
\end{tabular}
\end{table}

\subsubsection{Baselines}
\par We compare generative CDMs with two categoreis of baselines. I) \textbf{Transductive cognitive diagnosis models} for score prediction. The baseline includes DINA, IRT, MIRT and NCDM. Transductive CDMs model learner and item features via score prediction. These models can predict response scores for existing learners and items, but \textit{cannot instantly diagnose new learners' cognitive stats and reconstruct response scores}. For the fairness of comparison, we further choose II) \textbf{Encoder-decoder user modeling methods} for score reconstruction. The baseline includes U-AutoRec and CDAE.

\subsection{Experiment Analysis}


\subsubsection{Score Reconstruction Performance (RQ1)}
\par Table \ref{tab:score-reconstruction} presents the score reconstruction performance results for RQ1, comparing our proposed generative cognitive diagnosis models against traditional encoder-decoder baselines on two real-world datasets. The results demonstrate the superior performance of the generative paradigm across multiple evaluation metrics. On the ASSIST dataset, both G-IRT and G-NCDM achieve substantial improvements over the encoder-decoder methods, with G-NCDM reaching the highest accuracy (0.735) and best RMSE (0.433), while G-IRT attains the best F1-Score (0.827). Notably, even G-IRT without training achieves competitive F1-Score performance (0.816), highlighting the inherent effectiveness of the generative modeling approach. On the Math 1 dataset, G-IRT demonstrates consistent superiority with the highest accuracy (0.782) and lowest RMSE (0.408), while G-NCDM shows comparable performance to the best encoder-decoder baseline CDAE. The performance gains are particularly pronounced on the ASSIST dataset, where generative models achieve accuracy improvements of approximately 3-4\% and F1-Score improvements of 2-3\% over traditional methods. These results validate that the generative paradigm effectively addresses the score reconstruction task while maintaining the capability for instant diagnosis of new learners without requiring parameter re-optimization, thus confirming the practical advantages of our proposed approach.

\begin{figure}[htp]
    \subfloat[ASSIST ($n=832$)]{\includegraphics[width=0.5\linewidth]{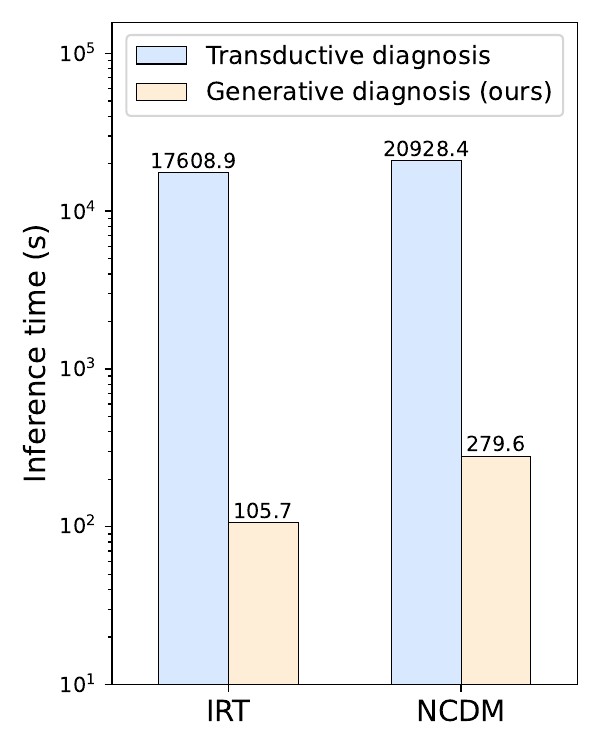}}
    \subfloat[Math 1 ($n=842$)]{\includegraphics[width=0.5\linewidth]{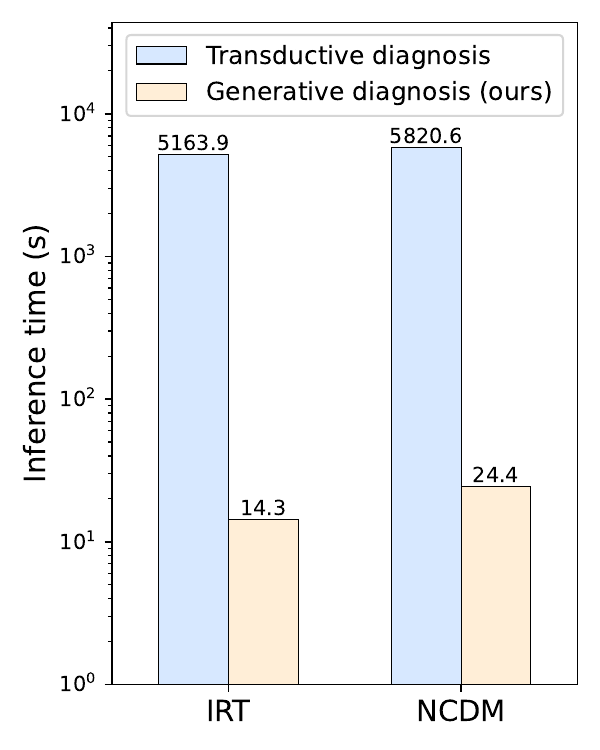}}
    \caption{Total running time for diagnosis of new learners using transductive and generative CDMs. Y-ticks are in the logarithmic scale for convenience. The $n$ denotes the number of new learners.}
    \label{fig:inference-time}
\end{figure}

\begin{table*}[htp]
\centering
\caption{Score reconstruction performance of generative user / learner modeling (RQ1).}\label{tab:score-reconstruction}
\begin{tabular}{@{}llllllll@{}}
\toprule
\multirow{2}{*}{Paradigm} & \multirow{2}{*}{Model} & \multicolumn{3}{c}{ASSIST (User Split)} & \multicolumn{3}{c}{Math 1 (User Split)} \\ \cmidrule(l){3-5} \cmidrule(l){6-8}
 &  & ACC$\uparrow$ & F1-Score$\uparrow$ & RMSE$\downarrow$ & ACC$\uparrow$ & F1-Score$\uparrow$ & RMSE$\downarrow$ \\ \cmidrule(r){1-8}
\multirow{2}{*}{\begin{tabular}[c]{@{}l@{}}Encoder\\ -decoder\end{tabular}} & U-AutoRec & 0.707\tiny$\pm$0.001 & 0.793\tiny$\pm$0.001 & \underline{0.438\tiny$\pm$0.001} & 0.749\tiny$\pm$0.00x & 0.728\tiny$\pm$0.00x & 0.419\tiny$\pm$0.00x \\
 & CDAE & 0.701\tiny$\pm$0.001 & 0.796\tiny$\pm$0.001 & 0.449\tiny$\pm$0.001 & 0.758\tiny$\pm$0.00x & \underline{0.746\tiny$\pm$0.00x} & 0.425\tiny$\pm$0.00x \\
 \midrule
\multirow{3}{*}{Generative CD} & G-IRT w/o train & 0.707\tiny$\pm$0.001 & 0.816\tiny$\pm$0.001 & 0.533\tiny$\pm$0.002 & \underline{0.781\tiny$\pm$0.00x} & 0.728\tiny$\pm$0.00x & \underline{0.411\tiny$\pm$0.00x} \\
 & G-IRT & \underline{0.734\tiny$\pm$0.001} & \textbf{0.827\tiny$\pm$0.001} & 0.451\tiny$\pm$0.001 & \textbf{0.782\tiny$\pm$0.00x} & 0.731\tiny$\pm$0.00x & \textbf{0.408\tiny$\pm$0.00x}\\
 & G-NCDM & \textbf{0.735\tiny$\pm$0.001} & \underline{0.822\tiny$\pm$0.001} & \textbf{0.433\tiny$\pm$0.001} & 0.749\tiny$\pm$0.00x & \textbf{0.747\tiny$\pm$0.00x} & 0.420\tiny$\pm$0.00x \\ \bottomrule
\end{tabular}
\end{table*}

\begin{table*}[htp]
\centering
\caption{Score prediction performance of transductive / generative cognitive diagnosis (RQ2).}
\begin{tabular}{@{}llllllll@{}}
\toprule
\multirow{2}{*}{Paradigm} & \multirow{2}{*}{Model} & \multicolumn{3}{c}{ASSIST (Random Split)} & \multicolumn{3}{c}{Math 1 (Random Split)} \\ \cmidrule(l){3-5} \cmidrule(l){6-8}
 &  & ACC$\uparrow$ & F1-Score$\uparrow$ & RMSE$\downarrow$ & ACC$\uparrow$ & F1-Score$\uparrow$ & RMSE$\downarrow$ \\ \cmidrule(r){1-8}
\multirow{5}{*}{Transductive CD} & DINA & 0.665\tiny$\pm$0.001 & 0.483\tiny$\pm$0.001 & 0.800\tiny$\pm$0.002 & 0.588\tiny$\pm$0.001 & 0.682\tiny$\pm$0.001  & 0.475\tiny$\pm$0.001 \\
 & IRT & 0.673\tiny$\pm$0.002 & 0.792\tiny$\pm$0.001 & 0.462\tiny$\pm$0.002 & 0.703\tiny$\pm$0.002  & \underline{0.702\tiny$\pm$0.001} & 0.443\tiny$\pm$0.001 \\
 & MIRT & 0.697\tiny$\pm$0.002 & 0.773\tiny$\pm$0.001 & 0.472\tiny$\pm$0.001 & 0.708\tiny$\pm$0.001 & 0.667\tiny$\pm$0.001 & 0.440\tiny$\pm$0.001 \\
 & NCDM & \underline{0.720\tiny$\pm$0.008} & 0.792\tiny$\pm$0.001 & \underline{0.433\tiny$\pm$0.002} & 0.727\tiny$\pm$0.001 & 0.668\tiny$\pm$0.002 & \underline{0.416\tiny$\pm$0.001} \\
 \midrule
\multirow{3}{*}{Generative CD} & G-IRT w/o train & 0.700\tiny$\pm$0.001 & \underline{0.810\tiny$\pm$0.001} & 0.530\tiny$\pm$0.001 & \underline{0.731\tiny$\pm$0.001} & 0.649\tiny$\pm$0.001 & 0.427\tiny$\pm$0.001 \\
 & G-IRT & 0.719\tiny$\pm$0.001 & 0.800\tiny$\pm$0.001 & 0.440\tiny$\pm$0.001 & \textbf{0.734\tiny$\pm$0.001} & 0.657\tiny$\pm$0.001 & 0.425\tiny$\pm$0.001\\
 & G-NCDM & \textbf{0.734\tiny$\pm$0.002} & \textbf{0.811\tiny$\pm$0.001} & \textbf{0.428\tiny$\pm$0.002} & \textbf{0.734\tiny$\pm$0.001} & \textbf{0.710\tiny$\pm$0.001} & \textbf{0.413\tiny$\pm$0.001}\\ \bottomrule
\end{tabular}
\end{table*}

\subsubsection{Score Prediction Performance (RQ2)}
\par Table II presents the score prediction performance results for RQ2, evaluating both generative and transductive cognitive diagnosis models on existing learners using random data splits. The results reveal that our proposed generative models achieve remarkable performance, often surpassing traditional transductive methods that were specifically designed for offline diagnosis scenarios. On the ASSIST dataset, G-NCDM demonstrates superior performance across all metrics, achieving the highest accuracy (0.734), F1-Score (0.811), and lowest RMSE (0.428), outperforming the best transductive baseline NCDM by 1.4\% in accuracy and 1.9\% in F1-Score. Similarly, on the Math 1 dataset, both G-IRT and G-NCDM achieve identical and superior accuracy (0.734) compared to the best transductive method NCDM (0.727), with G-NCDM showing the most balanced performance with the highest F1-Score (0.710) and lowest RMSE (0.413). Notably, even G-IRT without training demonstrates competitive performance, particularly on the ASSIST dataset where it achieves 0.700 accuracy and 0.810 F1-Score, highlighting the inherent strength of the generative modeling paradigm. These results are particularly significant as they demonstrate that generative models not only excel at instant diagnosis for new learners (as shown in RQ1) but also maintain competitive or superior performance in traditional offline diagnosis scenarios, effectively bridging the gap between inductive and transductive learning paradigms in cognitive diagnosis.

\begin{table*}[t]\centering
  \caption{Identifiability Score (IDS $\uparrow$) of diagnostic results of CDMs (RQ3). $\mathcal{I}(X)$ indicates whether $X$ is identifiable.}\label{tab:ids}
  \begin{tabular}{llllcllc}
    \toprule
    \multirow{2}{*}{Paradigm} & \multirow{2}{*}{Model} & \multicolumn{3}{c}{IDS $\uparrow$ of Learner Diagnostic Result $\Theta$} & \multicolumn{3}{c}{IDS $\uparrow$ of Question Diagnostic Result $\Psi$} \\
    \cmidrule(lr){3-5}\cmidrule(lr){6-8}
     & & ASSIST & Math1 & $\mathcal{I}(\Theta)$ & ASSIST & Math1 & $\mathcal{I}(\Psi)$\\
    \midrule
    \multirow{5}{*}{Transductive CD} & DINA & 0.550\tiny$\pm$0.003 & 0.451\tiny$\pm$0.006 & \XSolidBrush & 0.208\tiny$\pm$0.001 &  0.193\tiny$\pm$0.019 & \XSolidBrush \\
    & IRT & 0.691\tiny$\pm$0.004 & 0.690\tiny$\pm$0.004 & \XSolidBrush & 0.376\tiny$\pm$0.001 & 0.543\tiny$\pm$0.035 & \XSolidBrush \\
    & MIRT & 0.047\tiny$\pm$0.001 & 0.046\tiny$\pm$0.000 & \XSolidBrush & 0.041\tiny$\pm$0.000 & 0.085\tiny$\pm$0.005 & \XSolidBrush \\
    & NCDM & 0.857\tiny$\pm$0.001 & 0.662\tiny$\pm$0.005 & \XSolidBrush & 0.616\tiny$\pm$0.000 & 0.420\tiny$\pm$0.012 & \XSolidBrush \\
    & NCDM-Const & 0.897\tiny$\pm$0.001 & 0.688\tiny$\pm$0.003 & \XSolidBrush & 0.968\tiny$\pm$0.000 & 0.915\tiny$\pm$0.010 & \XSolidBrush \\
    \midrule
    \multirow{2}{*}{Generative CD} & G-IRT & \textbf{1.000\tiny$\pm$0.000}  & \textbf{1.000\tiny$\pm$0.000} & \CheckmarkBold & \textbf{1.000\tiny$\pm$0.000} & \textbf{1.000\tiny$\pm$0.000} & \CheckmarkBold \\
    & G-NCDM & \textbf{1.000\tiny$\pm$0.000}  & \textbf{1.000\tiny$\pm$0.000} & \CheckmarkBold & \textbf{1.000\tiny$\pm$0.000} & \textbf{1.000\tiny$\pm$0.000} & \CheckmarkBold \\
    \bottomrule
  \end{tabular}
\end{table*}

\begin{figure}[t]
    \centering
    \includegraphics[width=0.8\linewidth]{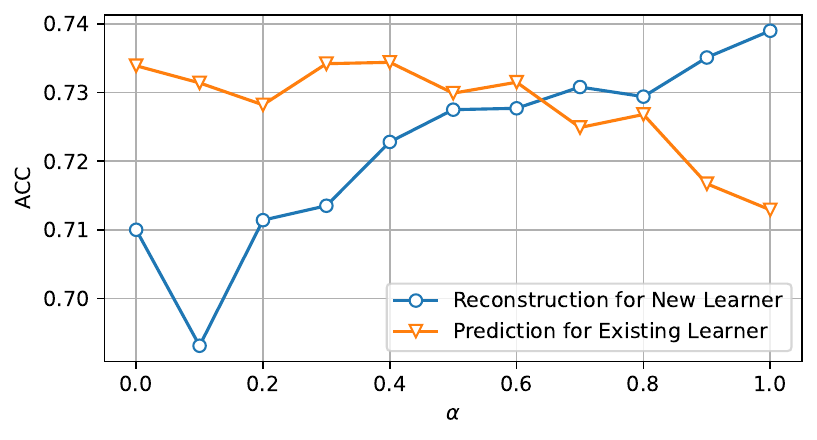}
    \caption{Score reconstruction \& prediction performance of G-NCDM w.r.t. hyperparameter $\alpha$ in ASSIST. The higher the $\alpha$, the large the weight of $\bm{\theta}_i^{(\text{exp})}$ in generating $\bm{\theta}_i$.}
    \label{fig:alpha-acc}
\end{figure}

\begin{figure}[t]
    \subfloat[Overall abilities.]{\includegraphics[width=0.5\linewidth]{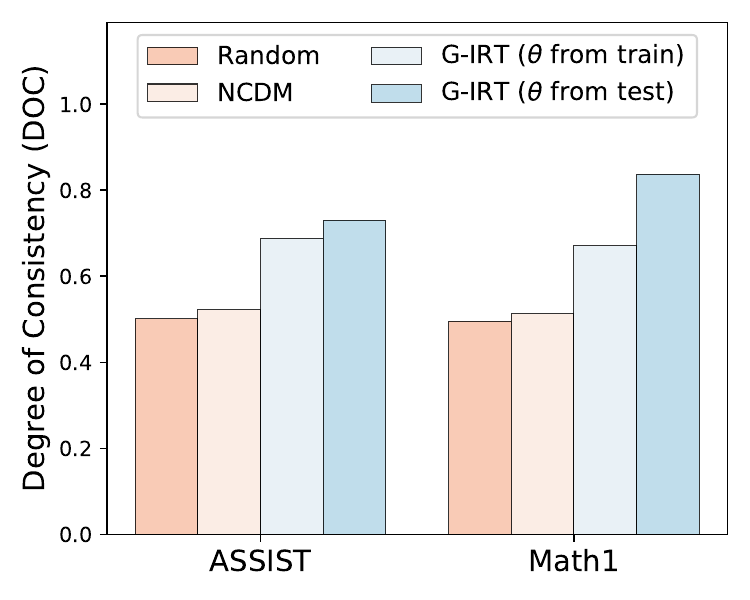}}
    \subfloat[Knowledge proficiencies.]{\includegraphics[width=0.5\linewidth]{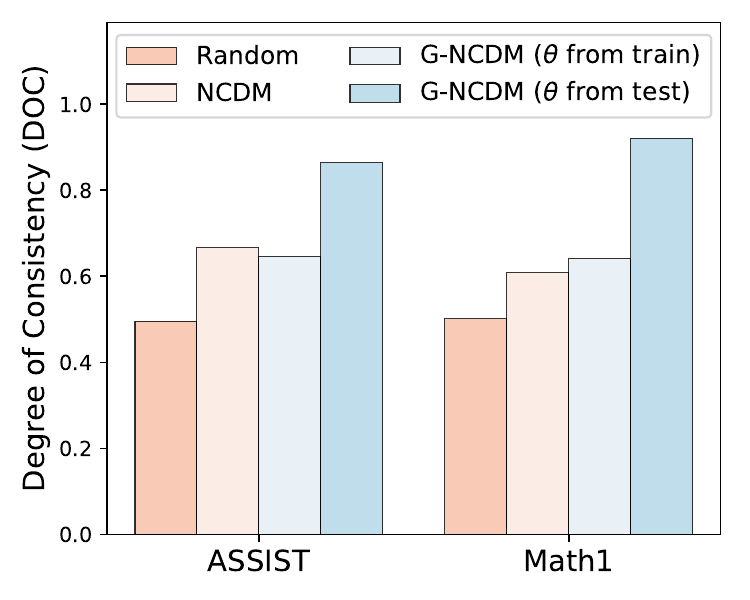}}
    \caption{Degree of consistency (DOC$\uparrow$) of diagnsotic results (RQ3).}
    \label{fig:doc}
\end{figure}

\subsubsection{Reliability of Diagnostic Outputs (RQ3)}
\par We analyze the reliabiity of diagnostic outputs from their identifiability and explainability.
\par \textbf{Identifiability of diagnostic results}. to quantitatively validate the identifiability of CDMs on the augmented data, we propose the \textit{Identifiability Score} (IDS) as an indicator of the identifiability. The more similar diagnostic results of original and shadow learners/questions, the larger the value of IDS. In addition, the full score of IDS denotes rigorous identifiability. To achieve this goal, we define IDS of learner traits $\Theta$ as follows:
\begin{equation}\label{eq:ids}
  IDS(\Theta) = \frac{1}{Z}\sum_{i \in S}\sum_{j \in S}\frac{I(\bm{r}_{i}=\bm{r}_j) \land I(i \neq j)}{\left[1 + dist(\bm{\theta}_i, \bm{\theta}_j)\right]^2},
\end{equation}
where $Z = \sum_{i \in S}\sum_{j \in S} I(\bm{r}_{i}=\bm{r}_j) \land I(i \neq j)$. The $dist(\bm{\theta}_i, \bm{\theta}_j)$ is the Manhattan distance \cite{Craw2010} between learner $i$'s traits and learner $j$'s traits. As mentioned above, $IDS(\Theta)$ is monotonically decreasing at $dist(\bm{\theta}_i, \bm{\theta}_j)$. \textbf{Learner traits are rigorously identifiable if and only if} $IDS(\Theta) = 1$. Similarly, we can also evaluate the identifiability of question parameters $\Psi$ by calculating $IDS(\Psi)$. Table \ref{tab:ids} presents the identifiability analysis results for RQ3, revealing a fundamental advantage of generative cognitive diagnosis models in terms of diagnostic reliability. The results demonstrate a stark contrast between traditional transductive and our proposed generative approaches in terms of identifiability guarantees. All traditional transductive methods (DINA, IRT, MIRT, NCDM, and NCDM-Const) fail to achieve identifiability for both learner diagnostic results ($\Theta$) and question diagnostic results ($\Psi$), as indicated by the ``\XSolidBrush'' symbols, with their Identifiability Scores (IDS) ranging from as low as 0.041 (MIRT) to 0.968 (NCDM-Const) across different datasets. In sharp contrast, both G-IRT and G-NCDM achieve perfect identifiability with IDS scores of 1.000 for both learner and question diagnostics on both ASSIST and Math1 datasets, as confirmed by the ``\CheckmarkBold'' symbols. This perfect identifiability ensures that the diagnostic outputs produced by our generative models are theoretically guaranteed to be unique and reliable, addressing a critical limitation of traditional cognitive diagnosis methods. The superior identifiability performance of generative models stems from their well-designed generation process that explicitly incorporates identifiability conditions, providing practitioners with diagnostic results they can trust for high-stakes educational assessment and decision-making scenarios.

\begin{figure*}[htp]
    \centering
    \subfloat[IRT@ASSIST]{\includegraphics[width=0.33\linewidth]{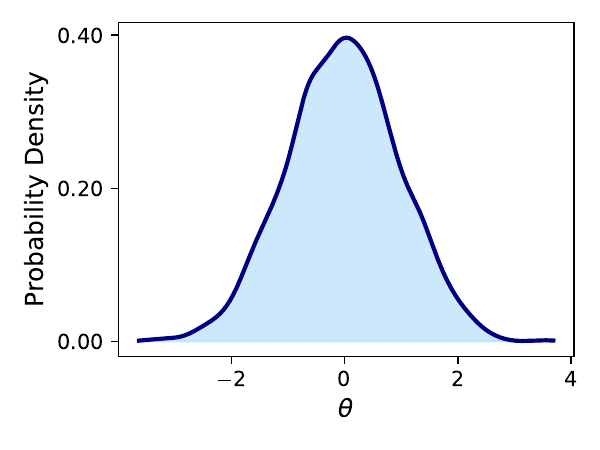}}
    \subfloat[G-IRT@ASSIST]{\includegraphics[width=0.33\linewidth]{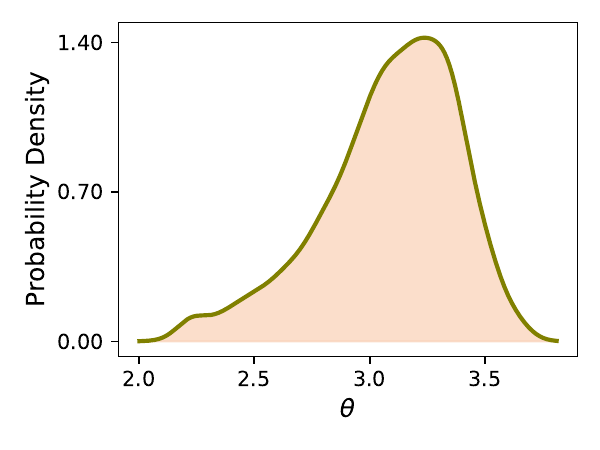}}
    \subfloat[Correct@ASSIST]{\includegraphics[width=0.33\linewidth]{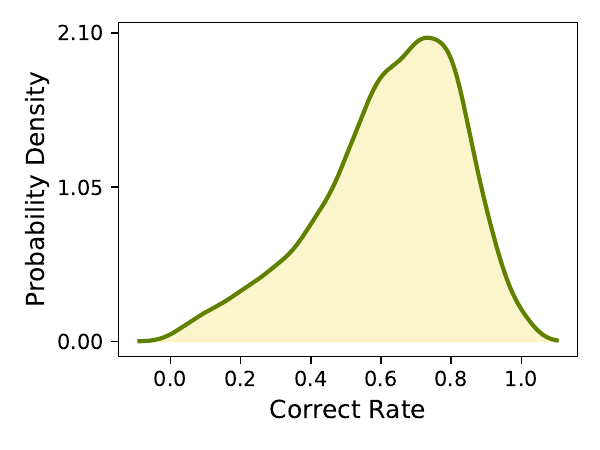}}
    \caption{Visualization of diagnostic results using KDE. Learners w/o response logs are removed from the visualization.}
    \label{fig:irt-pdf-assist}
\end{figure*}

\begin{figure*}[htp]
    \centering
    \subfloat[IRT@Math 1]{\includegraphics[width=0.33\linewidth]{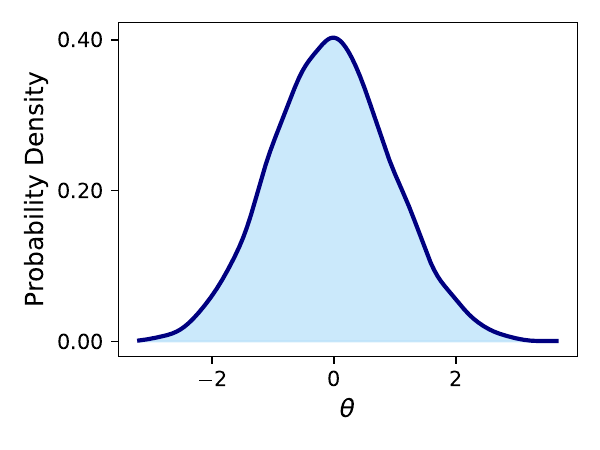}}
    \subfloat[G-IRT@Math 1]{\includegraphics[width=0.33\linewidth]{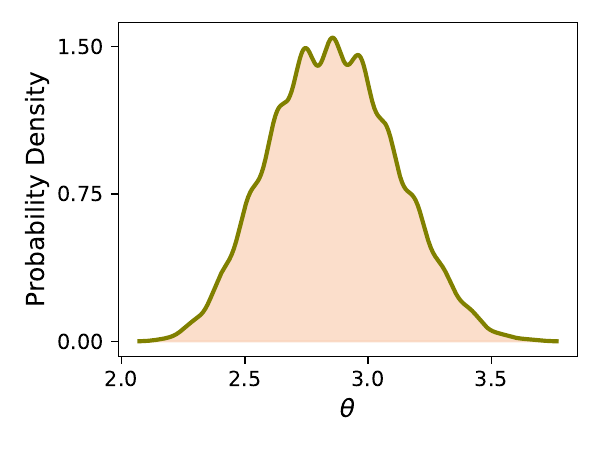}}
    \subfloat[Correct@Math 1]{\includegraphics[width=0.33\linewidth]{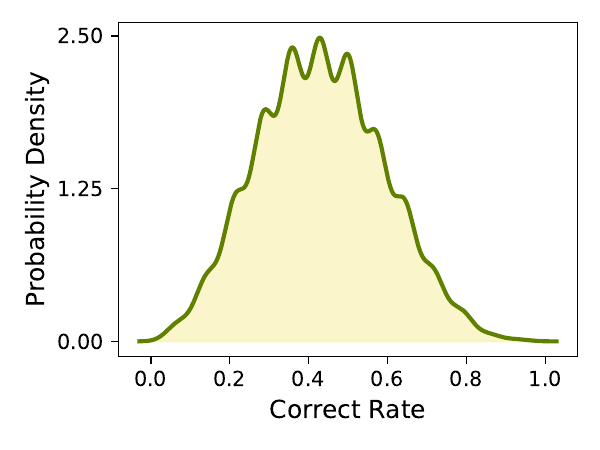}}
    \caption{Visualization of diagnostic results using KDE. Learners w/o response logs are removed from the visualization.}
    \label{fig:irt-pdf-math1}
\end{figure*}

\par\textbf{Explainability of diagnostic results}. Figure \ref{fig:doc} presents the explainability analysis results for RQ3. Our motivation is that the order of explainable learners' knowledge proficiencies should be consistent with the order of response scores on relevant questions. To this end, inspired by previous works \cite{FoussPRS2007}, we propose the \textit{Degree of Consistency} (DOC) as the evaluation metric. Given question $e_l, l=1,2,\ldots,M$, DOC is defined as follows:
\begin{equation}
DOC(e_l)\!=\!\frac{\sum_{i,j}\delta (r_{il},\! r_{jl})\sum_{k=1}^K q_{lk}\!\land\! J(l,i,j)\!\land\!\delta(\theta_{ik},\!\theta_{jk})}{\sum_{i,j}\delta (r_{il},\! r_{jl})\sum_{k=1}^K\! q_{lk}\!\land\! J(l,i,j)\!\land\! I(\theta_{ik}\!\neq\!\theta_{jk})},
\end{equation}
The DOC measures the monotonicity between diagnostic results and response scores for learners answering the same item. We present experiment results for IRTs and NCDMs respectively because the former diagnose learner cognitive states as overall ability (single-dimensional value), while the latter diagnose learner cognitive states as knowledge proficiencies (multi-dimensional value). Here, ``$\theta$ from train/test'' denotes that the evidence data for generative diagnosis is from the training/test dataset. Response scores for calculating DOC are from the test dataset. Regarding result analysis, we have two findings. First, it can be observed that the DOC of generative CDMs (i.e., G-IRT and G-NCDM) with $\theta$ from training dataset is mostly higher than that of the corresponding transductive CDMs. This observation indicates that generative CDMs can mostly capture the monotonicity between explicit response scores and implicit cognitive states, demonstrating the psychometrical explainability of diagnostic results for learners in various scenarios. Second, it can be observed that DOC of generative CDMs with $\theta$ from test dataset is always higher than other results. This essentially demonstrates the explainability of \textit{generative diagnosis}. As both the evidence data for diagnosing learner cognitive states and the response scores for calculating DOC are from the same dataset, generative CDMs can effectively capture the complex mapping between response scores and cognitive states, generating reliable diagnostic results while maintaining excellent score reconstruction accuracy (see Table \ref{tab:score-reconstruction}).

\begin{figure*}[htp]
    \centering
    \subfloat[NCDM@ASSIST]{\includegraphics[width=0.25\linewidth]{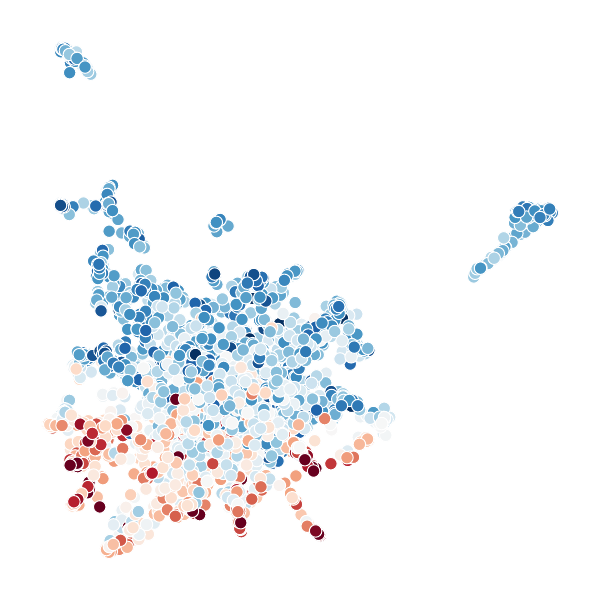}}
    \subfloat[G-NCDM@ASSIST]{\includegraphics[width=0.25\linewidth]{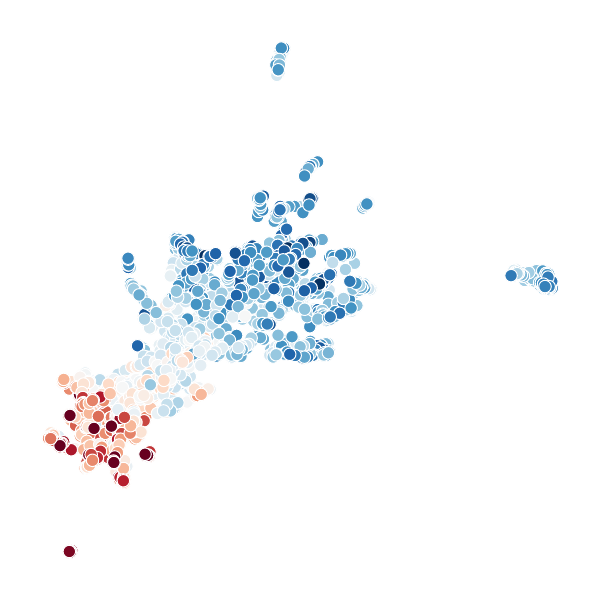}}
    \subfloat[NCDM@Math 1]{\includegraphics[width=0.25\linewidth]{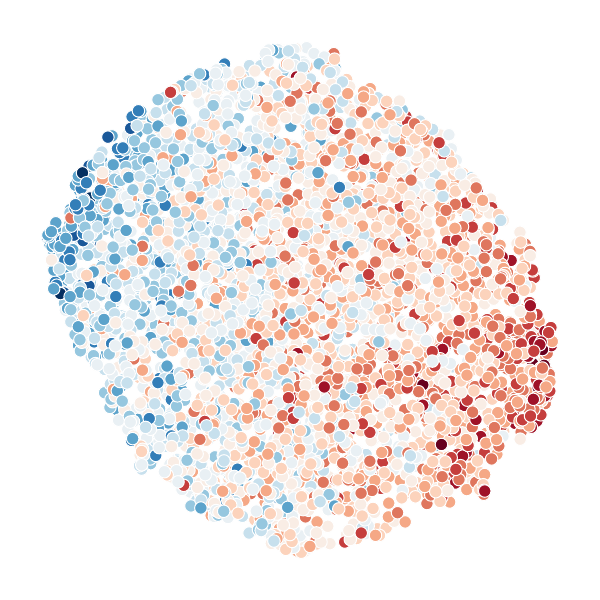}}
    \subfloat[G-NCDM@Math 1]{\includegraphics[width=0.25\linewidth]{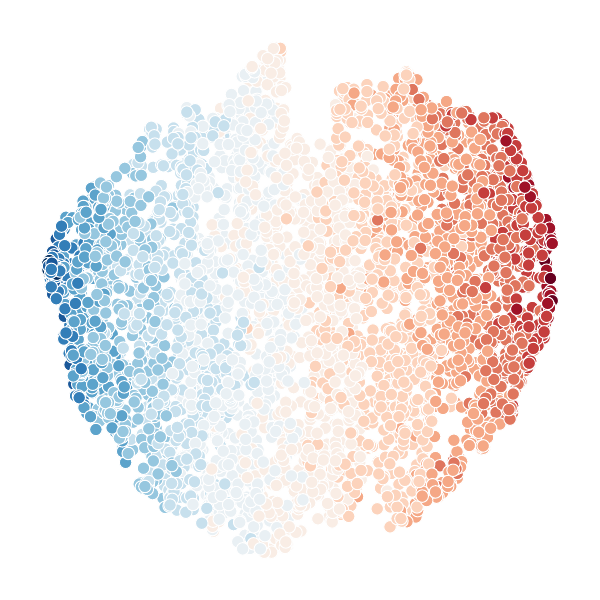}}
    \caption{Visualization of diagnostic results using u-map, colored with learner score rates. Learners w/o response logs are removed from the visualization.}
    \label{fig:umap-filtered}
\end{figure*}

\begin{figure}[t]
    \centering
    \subfloat[NCDM.]{\includegraphics[width=0.5\linewidth]{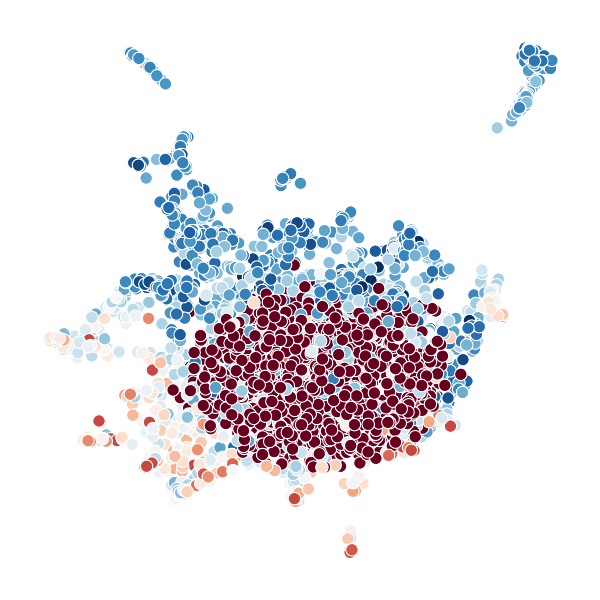}}
    \subfloat[G-NCDM.]{\includegraphics[width=0.5\linewidth]{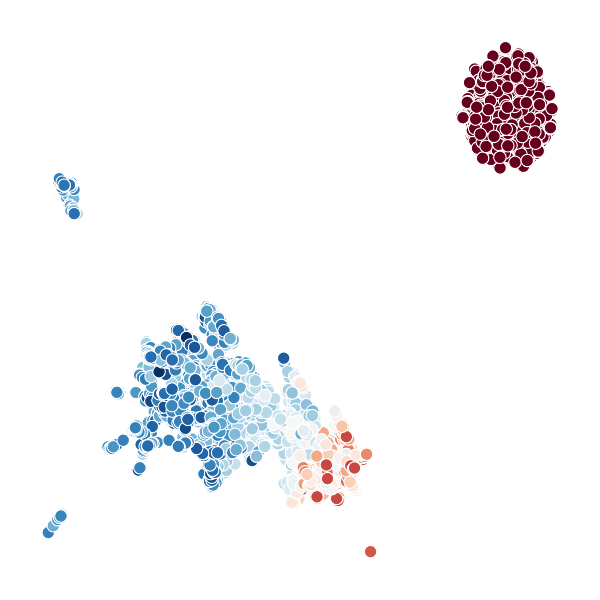}}
    \caption{Visualization of diagnostic results in ASSIST using u-map, colored with learner score rates. All learners in the raw data, including those without response logs (colored with \textbf{dark red}), are included in the visualization.}
    \label{fig:umap-raw}
\end{figure}

\subsubsection{Distribution Analysis of G-IRT (RQ4.1)} 
\par To explore the utility of G-IRT, we visualize the histogram of estimated learner ability $\theta$ and correct rates. Figure \ref{fig:irt-pdf-assist} and \ref{fig:irt-pdf-math1} present the visualization results, with the probability density estimated by the kernel density estimation [cite]. The evaluation motivation is that the distribution of appropriate learner diagnosis results should present association with the distribution of observed learner correct rates. From figure \ref{fig:irt-pdf-assist}, we can observe that the distribution of learner correct rates is skewed. This property is well preserved by the distribution of diagnostic results of G-IRT. From figure \ref{fig:irt-pdf-math1}, we Tcan observe that the distribution of learner correct rates is multimodal. This property is also well preserved by G-IRT. These observations demonstrate the advantage of the generative diagnosis paradigm. The reason is that, according to Eq. \ref{eq:girt-theta-single} and Eq. \ref{eq:girt-theta-overall}, the generative process of learner ability $\theta$ is equivalent to calculating an weighted average of response scores. Therefore, the calculation of score rate can be viewed as a special case of the generative diagnosis process. On the contrary, transductive IRT relies on parameter estimation to diagnose learner ability, which loses the information of the distribution of overall correct rate during parameter estimation of learner ability.

\subsubsection{Clustering Analysis of G-NCDM (RQ4.2)}
\par To explore the statistical relationship between diagnosed learner traits and learners' actual performance, we visualize learner diagnostic results of CDMs. For NCDM and G-NCDM that model learner cognitive states by multi-dimensional knowledge proficiencies, we visualize their knowledge proficiencies by UMAP \cite{mcinnes2020umap} and color points of learners by their correct rates. Then we explore whether diagnostic results can distinguish between learners with different correct rates. Figure \ref{fig:umap-filtered} and Figure \ref{fig:umap-raw} presents the visualization results. We obtain two findings among them:

\par \textbf{Finding 1. cognitive states diagnosed by G-NCDM is better clustered w.r.t. correct rates.} Figure \ref{fig:umap-filtered} presents visualization results that has removed learners without response scores, as similar as in previous studies [cite]. From the visualization on ASSIST, we can observe that cognitive states generated by G-NCDM is linear separable, as red points (i.e., learners with correct rate < 0.5) aggregates in the bottom left of the figure, while blue points (i.e., learners with correct rate > 0.5) spread in the center of the figure. From the visualization on Math 1, we can also observed that points with similar color (i.e., learners with similar correct rate) aggregates tighter for G-NCDM@Math 1, while points with different colors usually mix together within a small region for NCDM@Math 1. Both the linear separatability and the tight aggregation of similar color demonstrate the statistical reliability of G-NCDM.

\par \textbf{Finding 2. G-NCDM can effectivelly recognize empty learners.} Online learning platforms inevitably include empty learners that have not any response score, like the ASSIST dataset introduced in Table \ref{tab:dataset-overview}. We present in Figure \ref{fig:umap-raw} cognitive states of all learners including those empty. It could be observed that empty learners (dark red points) are well separated and aggregated in the top right of the visualization of G-NCDM, while they overlap with other learners in the visualization of NCDM. Essentially, this ability of G-NCDM originates from the generative diagnosis mechanism. Since the score vectors of all empty learners is the zero vector, their diagnostic results are also the same and different from other learners. This finding validates the generalization ability and outlier detection ability of G-NCDM.

\begin{table*}[t]
    \centering
    \caption{Raw response score cases of learners in Math 1(RQ5). Each column denotes response scores on an item.}\label{tab:raw-responses}
    \begin{tabular}{ccccccccccccccccccccc}
    \toprule
        Learner & 1 & 2 & 3 & 4 & 5 & 6 & 7 & 8 & 9 & 10 & 11 & 12 & 13 & 14 & 15 & 16 & 17 & 18 & 19 & 20\\
        \midrule
        1737 & \CheckmarkBold & \CheckmarkBold & \CheckmarkBold & \CheckmarkBold & \XSolidBrush & \XSolidBrush & \CheckmarkBold & \CheckmarkBold & \CheckmarkBold & \XSolidBrush & \CheckmarkBold & \CheckmarkBold & \XSolidBrush & \CheckmarkBold & \XSolidBrush & \CheckmarkBold & \XSolidBrush & \XSolidBrush & \XSolidBrush & \XSolidBrush \\
        2094 & \CheckmarkBold & \XSolidBrush & \XSolidBrush & \CheckmarkBold & \XSolidBrush & \XSolidBrush & \CheckmarkBold & \CheckmarkBold & \XSolidBrush & \XSolidBrush & \XSolidBrush & \XSolidBrush & \XSolidBrush & \CheckmarkBold & \XSolidBrush & \XSolidBrush & \CheckmarkBold & \XSolidBrush & \XSolidBrush & \XSolidBrush \\
         \bottomrule
    \end{tabular}
\end{table*}

\begin{table*}[t]
    \centering
    \caption{Knowledge correct rates in Math 1(RQ5). Each column denotes average correct rates on items requiring the knowledge.}\label{tab:know-correct}
    \begin{tabular}{cccccccccccc}
    \toprule
        Learner & A1 & A2 & A3 & A4 & A5 & A6 & A7 & A8 & A9 & A10 & A11 \\
        \midrule
        1737 & 0.67 & 1.0 & 0.43 & 1.0 & 0.75 & 0.33 & 0.0 & 0.40 & 0.50 & 0.42 & 0.59 \\
        2094 & 0.33 & 1.0 & 0.0 & 0.0 & 1.0 & 0.17 & 0.0 & 0.0 & 0.50 & 0.42 & 0.35 \\
        \bottomrule
    \end{tabular}
\end{table*}

\begin{figure*}[t]
    \centering
    \includegraphics[width=\linewidth]{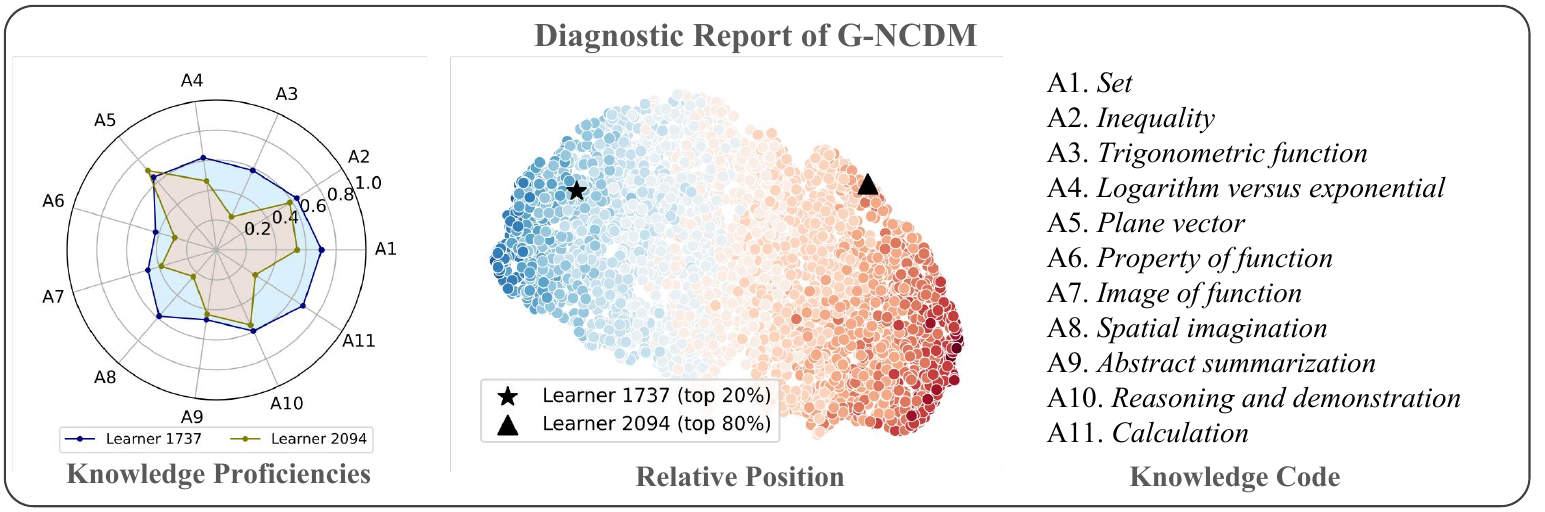}
    \caption{A case of diagnostic report generated by G-NCDM. The middle term shows the visualization of all learners' diagnostic outputs using u-map, with each point colored by the learner's correct rate (blue denotes higher correct rates while red denotes lower correct rates).}
    \label{fig:case-gncdm}
\end{figure*}

\begin{figure}[t]
    \centering
    \includegraphics[width=0.9\linewidth]{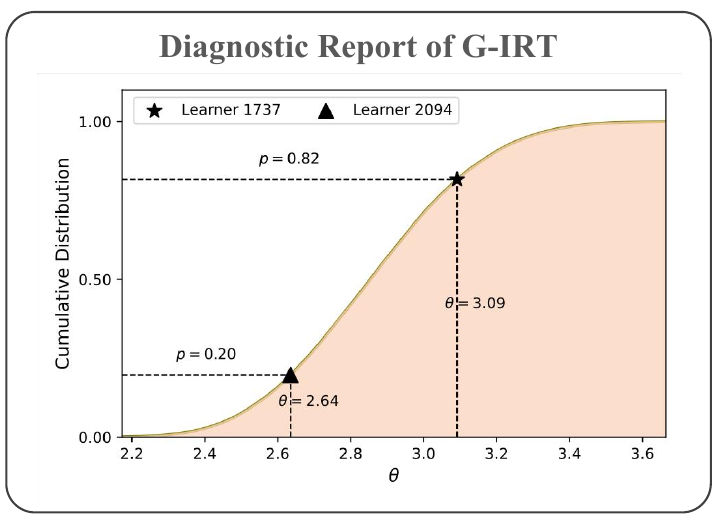}
    \caption{A case of diagnostic report generative by G-IRT. The orange curve presents the cumulative distribution function (CDF) of generated learner ability. The term $p$ denotes the CDF value of learners, which is equivalent to the ratio of learners whose correct rates are lower than the target learner.}
    \label{fig:case-girt}
\end{figure}

\subsubsection{A Case Study of G-IRT and G-NCDM (RQ5)}
\par We explore the utility of generative cognitive diagnosis models in real-world scenarios via a case study. Specifically, we select Math 1 for the case study because the dataset comes from the representative offline test scenario and the response scores are dense. We then select two learners, including one ranked top 20\% and one ranked top 80\% by correct rates. We present their raw response scores and knowledge correct rates, as shown in Table \ref{tab:raw-responses} and \ref{tab:know-correct}. We then input their data into G-IRT and G-NCDM respectively to obtain and visualize the diagnostic reports. Each diagnostic report consisting two components: \textbf{1) the learner's cognitive state} and \textbf{2) the learner's relative position in the learner group}. We make a crossover comparison between the two learners' response data and diagnostic reports to see if G-IRT and G-NCDM works for this scenario. According to the case study, we obtain two findings.
\par \textbf{Finding 1. G-NCDM can well reflect the real knowledge proficiency of learners.} By comparing observed knowledge correct rates (see Table \ref{tab:know-correct}) and generated knowledge proficiencies (see the left of Figure \ref{fig:case-gncdm}), we find that the latter is consistent with the former. For instance, for learner 1737 and 2094, only the knowledge correct rate on A5 ``\textit{plane vector}'' of the former is lower than that of the latter. This is exactly reflected in their generated knowledge proficiencies. In addition, both learners perform poor on knowledge $A6$ ``\textit{property of function}'' and A7 ``\textit{image of function}'', with correct rates less than 0.33. This is also well reflected in their knowledge proficiencies, with values always less than 0.4. These observations essentially demonstrate the effectiveness of the forked design of the generative diagnosis function of G-NCDM (see Eq. \ref{eq:gncdm-gdf}). Technically, the direct, parameter-free generation of $\bm{\theta}_i^{(\text{exp})}$ ensures that diagnostic results largely reflect knowledge correct rates, while the neural network-based generation of $\bm{\theta}_i^{(\text{imp})}$ calibrates diagnostic results for more fine-grained diagnosis and better score prediction performance (see Figure \ref{fig:alpha-acc}). The center of Figure \ref{fig:case-gncdm} presents the learners' relative position in the learner group, which also reflect their different cognitive states. Overall, G-NCDM can well reflect knowledge concept-wise cognitive states in this scenario.
\par \textbf{Finding 2. G-IRT is beneficial to learner ranking and overall cognitive state diagnosis.} As shown in Figure \ref{fig:case-girt}, the diagnostic report of G-IRT is presented as a cumulative distribution function of generated learer abilities. We can observe from the report that learner 1737 is ranked top $(1-p)\times 100\% = 17\%$ ($\theta = 3.09$), and learner 2094 is ranked top $(1-p)\times 100\% = 78\%$ ($\theta = 2.64$). These results are consistent with observed response data, demonstrating the effectiveness of G-IRT in real-world scenarios.

\section{Conclusion}
\par In this study, we proposed the generative cognitive diagnosis paradigm for the cognitive diagnosis task. The generative diagnosis paradigm overcomes two significant challenges of traditional transductive cognitive diagnosis models (CDMs). The first challenge is their unavailability for instant diagnosis of new-coming learners. Cognitive diagnosis for new-coming learners requires retraining transductive CDMs, which is resource-consuming and can lead to the inconsistency for existing learner cognitive states. The second challenge is a lack of reliability of diagnostic outputs. Diagnostic outputs provided by transductive CDMs are not indentifiable and lack psychometric reliability, which orginates from the intrinsic properties of parameter optimization algorithms. The generative diagnosis paradigm effectively overcomes the challenges via the generative diagnosis procedure, which generates learner cognitive states using a generative diagnosis function (GDF) rather than parameter optimization. Technically, we further proposed two instantiations of the generative diagnosis paradigm, the Generative Item Response Theory (G-IRT) and the Generative Neural Cognitive Diagnosis Model (G-NCDM). Respecting G-IRT, we established the GDF by substituting unobservable parameters in the inverse of the item response function of IRT with ``proxy parameters''. We then analyzed mathematical properties of G-IRT and provided parameter constraints for better cold-start performance and controllability. Regarding G-NCDM, we propose a neural network-driven GDF based on the previous ID-CDM. Notably, G-NCDM introduced Q-matrix mapping in the generative process, which enhanced the relationship between diagnostic outputs and knowledge dimensions. Next, we demonstrated the effectiveness of generative CDMs via extensive experiments on two real-world cognitive diagnosis datasets. Experiment results revealed that generative CDMs not only has excellent score reconstruction (for instant diagnosis) and score prediction (for offline diagnosis) performances, but can generate reliable diagnostic outputs. These results demonstrate the accuracy, reliability and utility of generative CDMs in real-world applications of cognitive diagnosis.


\section{Discussion}
\par As a new paradigm of cognitive diagnosis, there are still many theoretical and technological limitations of generative cognitive diagnosis underexplored. Here we figure out limitations and future works of the generative cognitive diagnosis.

\subsection{Continual Learning}
\par The first limitation of this study is a lack of the continual learning ability of the proposed methods. Specifically, response data, test items and learners would continually accumulate in a learning platform, which requires the CDM to continually learn from new-coming data and update its diagnosis ability. Although current generative CDMs can instantly diagnose for new-coming learners, response data of new-coming learners and items still cannot be used for model updating. Along this line, we hope to develop mechanisms or algorithms that enable generative CDMs to continually learn from new-coming response data. Continual learning ability of generative CDMs is beneficial for a host of online learning / evaluation applications, such as student cognitive ability tests and large language model evaluation.

\subsection{Multi-modal Cognitive Diagnosis}
\par The second limitation of this study is a lack of multi-modal data modeling ability. Specifically, two types of multi-modal data a usually available in real-world cognitive ability tests, including \textbf{response behavioral data} and \textbf{item question data}. Response behavioral data, such as response time length, mouse click record and number of response attempts, contains rich information of learners' congitive state. However, behavioral data is resource-consuming to collect and rare in public cognitive diagnosis datasets. Item question data contains original question texts and images/videos. Recent advancements in LLM evaluation benchmarks, such as GSM8K and MMMU, provide abundent item question data. However, the lack of response scores on these data limits their utility in cognitive diagnosis. 

\subsection{Cognitive Diagnosis for Model Evaluation}
\par Generative cognitive diagnosis is suitable for intelligent model ability evaluation. Recent advancements of large language models boost the requirement for model evaluation. A significant advantage of cognitive diagnosis compared to score-driven evaluation metric is that CD decomposes model ability into abstract cognitive states (e.g., overall ability in G-IRT and knowledge proficiencies in G-NCDM), which helps predict model performance on items lacking their responses and deepen researcher's knowledge about model ability. As mentioned above, the instant diagnosis ability of generative cognitive diagnosis and the rich evaluation benchmarks enables reliable and efficient evaluation of LLMs. As an intelligent diagnosis tool, we believe that generative cognitive diagnosis could make essential contribution on the way to Artificial General Intelligence (AGI).

\bibliographystyle{ieeetr}
\bibliography{ref}

\begin{thebibliography}{10}

\bibitem{Torre2009}
J.~de~la Torre, ``Dina model and parameter estimation: A didactic,'' {\em Journal of Educational and Behavioral Statistics}, vol.~34, no.~1, pp.~115--130, 2009.

\bibitem{Fischer1995}
G.~H. Fischer, {\em Derivations of the Rasch Model}, pp.~15--38.
\newblock New York, NY: Springer New York, 1995.

\bibitem{Brzezinska2020}
J.~Brzezinska, ``Item response theory models in the measurement theory,'' {\em Commun. Stat. Simul. Comput.}, vol.~49, no.~12, pp.~3299--3313, 2020.

\bibitem{Gelfand1990}
A.~E. Gelfand and A.~F.~M. Smith, ``Sampling-based approaches to calculating marginal densities,'' {\em Journal of the American Statistical Association}, vol.~85, no.~410, pp.~398--409, 1990.

\bibitem{Hastings1970}
W.~K. Hastings, ``Monte carlo sampling methods using markov chains and their applications,'' {\em Biometrika}, vol.~57, no.~1, pp.~97--109, 1970.

\bibitem{Wu2020}
M.~Wu, R.~L. Davis, B.~W. Domingue, C.~Piech, and N.~D. Goodman, ``Variational item response theory: Fast, accurate, and expressive,'' in {\em {EDM}}, International Educational Data Mining Society, 2020.

\bibitem{Reckase2009}
M.~D. Reckase, {\em Multidimensional Item Response Theory Models}, pp.~79--112.
\newblock New York, NY: Springer New York, 2009.

\bibitem{Yeung2019}
C.~Yeung, ``Deep-irt: Make deep learning based knowledge tracing explainable using item response theory,'' in {\em {EDM}}, International Educational Data Mining Society {(IEDMS)}, 2019.

\bibitem{WangF2022}
F.~Wang, Q.~Liu, E.~Chen, Z.~Huang, Y.~Yin, S.~Wang, and Y.~Su, ``Neuralcd: A general framework for cognitive diagnosis,'' {\em IEEE Transactions on Knowledge and Data Engineering}, pp.~1--16, 2022.

\bibitem{Zhou0WWHT0CM2021}
Y.~Zhou, Q.~Liu, J.~Wu, F.~Wang, Z.~Huang, W.~Tong, H.~Xiong, E.~Chen, and J.~Ma, ``Modeling context-aware features for cognitive diagnosis in student learning,'' in {\em {KDD}}, pp.~2420--2428, {ACM}, 2021.

\bibitem{gao2021rcd}
W.~Gao, Q.~Liu, Z.~Huang, Y.~Yin, H.~Bi, M.~Wang, J.~Ma, S.~Wang, and Y.~Su, ``{RCD:} relation map driven cognitive diagnosis for intelligent education systems,'' in {\em {SIGIR} '21: The 44th International {ACM} {SIGIR} Conference on Research and Development in Information Retrieval, Virtual Event, Canada, July 11-15, 2021} (F.~Diaz, C.~Shah, T.~Suel, P.~Castells, R.~Jones, and T.~Sakai, eds.), pp.~501--510, {ACM}, 2021.

\bibitem{Li2022}
J.~Li, F.~Wang, Q.~Liu, M.~Zhu, W.~Huang, Z.~Huang, E.~Chen, Y.~Su, and S.~Wang, ``Hiercdf: {A} bayesian network-based hierarchical cognitive diagnosis framework,'' in {\em {KDD}}, pp.~904--913, {ACM}, 2022.

\bibitem{kingma2014vae}
D.~P. Kingma and M.~Welling, ``Auto-encoding variational bayes,'' in {\em 2nd International Conference on Learning Representations, {ICLR} 2014, Banff, AB, Canada, April 14-16, 2014, Conference Track Proceedings} (Y.~Bengio and Y.~LeCun, eds.), 2014.

\bibitem{goodfellow2020gan}
I.~Goodfellow, J.~Pouget-Abadie, M.~Mirza, B.~Xu, D.~Warde-Farley, S.~Ozair, A.~Courville, and Y.~Bengio, ``Generative adversarial networks,'' {\em Commun. ACM}, vol.~63, p.~139–144, Oct. 2020.

\bibitem{ho2020ddpm}
J.~Ho, A.~Jain, and P.~Abbeel, ``Denoising diffusion probabilistic models,'' in {\em Proceedings of the 34th International Conference on Neural Information Processing Systems}, NIPS '20, (Red Hook, NY, USA), Curran Associates Inc., 2020.

\bibitem{dhariwal2020diffusion}
P.~Dhariwal and A.~Nichol, ``Diffusion models beat gans on image synthesis,'' in {\em Proceedings of the 35th International Conference on Neural Information Processing Systems}, NIPS '21, (Red Hook, NY, USA), Curran Associates Inc., 2021.

\bibitem{Sutskever2014}
I.~Sutskever, O.~Vinyals, and Q.~V. Le, ``Sequence to sequence learning with neural networks,'' in {\em {NIPS}}, pp.~3104--3112, 2014.

\bibitem{devlin2019bert}
J.~Devlin, M.~Chang, K.~Lee, and K.~Toutanova, ``{BERT:} pre-training of deep bidirectional transformers for language understanding,'' in {\em Proceedings of the 2019 Conference of the North American Chapter of the Association for Computational Linguistics: Human Language Technologies, {NAACL-HLT} 2019, Minneapolis, MN, USA, June 2-7, 2019, Volume 1 (Long and Short Papers)} (J.~Burstein, C.~Doran, and T.~Solorio, eds.), pp.~4171--4186, Association for Computational Linguistics, 2019.

\bibitem{Sedhain2015}
S.~Sedhain, A.~K. Menon, S.~Sanner, and L.~Xie, ``Autorec: Autoencoders meet collaborative filtering,'' in {\em {WWW} (Companion Volume)}, pp.~111--112, {ACM}, 2015.

\bibitem{WuDZE2016}
Y.~Wu, C.~DuBois, A.~X. Zheng, and M.~Ester, ``Collaborative denoising auto-encoders for top-n recommender systems,'' in {\em {WSDM}}, pp.~153--162, {ACM}, 2016.

\bibitem{li2024idcdf}
J.~Li, Q.~Liu, F.~Wang, J.~Liu, Z.~Huang, F.~Yao, L.~Zhu, and Y.~Su, ``Towards the identifiability and explainability for personalized learner modeling: An inductive paradigm,'' in {\em {WWW}}, pp.~3420--3431, {ACM}, 2024.

\bibitem{Xu2018}
P.~Xu and M.~C. Desmarais, ``An empirical research on identifiability and q-matrix design for {DINA} model,'' in {\em {EDM}}, International Educational Data Mining Society {(IEDMS)}, 2018.

\bibitem{Xu2019identifiability}
G.~Xu, {\em Identifiability and Cognitive Diagnosis Models}, pp.~333--357.
\newblock Cham: Springer International Publishing, 2019.

\bibitem{Tatsuoka1983}
K.~K. Tatsuoka, ``Rule space: An approach for dealing with misconceptions based on item response theory,'' {\em Journal of Educational Measurement}, vol.~20, no.~4, pp.~345--354, 1983.

\bibitem{Feng2009}
M.~Feng, N.~T. Heffernan, and K.~R. Koedinger, ``Addressing the assessment challenge with an online system that tutors as it assesses,'' {\em User Model. User Adapt. Interact.}, vol.~19, no.~3, pp.~243--266, 2009.

\bibitem{Liu2018}
Q.~Liu, R.~Wu, E.~Chen, G.~Xu, Y.~Su, Z.~Chen, and G.~Hu, ``Fuzzy cognitive diagnosis for modelling examinee performance,'' {\em {ACM} Trans. Intell. Syst. Technol.}, vol.~9, no.~4, pp.~48:1--48:26, 2018.

\bibitem{Craw2010}
S.~Craw, {\em Manhattan Distance}, pp.~639--639.
\newblock Boston, MA: Springer US, 2010.

\bibitem{FoussPRS2007}
F.~Fouss, A.~Pirotte, J.~Renders, and M.~Saerens, ``Random-walk computation of similarities between nodes of a graph with application to collaborative recommendation,'' {\em {IEEE} Trans. Knowl. Data Eng.}, vol.~19, no.~3, pp.~355--369, 2007.

\bibitem{mcinnes2020umap}
L.~McInnes, J.~Healy, and J.~Melville, ``Umap: Uniform manifold approximation and projection for dimension reduction,'' 2020.

\end{thebibliography}

\end{document}